\documentclass[journal]{IEEEtran}

\usepackage{amsmath,amsfonts}
\usepackage{algorithmic}
\usepackage{algorithm}
\usepackage{array}
\usepackage{algorithmic}
\usepackage{amsmath}

\usepackage[caption=false,font=normalsize,labelfont=sf,textfont=sf]{subfig}
\usepackage{textcomp}
\usepackage{stfloats}
\usepackage{url}
\usepackage{verbatim}
\usepackage{graphicx}
\usepackage{cite}
\usepackage{enumerate}
\usepackage{amsmath}
\usepackage{color}
\usepackage{booktabs} 
\usepackage{array}
\usepackage{multirow}

\usepackage{lineno,hyperref}
\usepackage{hyperref}

\begin{document}
\title{Dual High-Order Total Variation Model for Underwater Image Restoration}

\author{Yuemei~Li,
Guojia~Hou,~\IEEEmembership{Member,~IEEE,}
Peixian~Zhuang,~\IEEEmembership{Senior Member,~IEEE,}
Zhenkuan\\Pan,~\IEEEmembership{Senior Member,~IEEE}
\vspace{-3em}
\thanks{This work has been submitted to the IEEE for possible publication. Copyright may be transferred without notice, after which this version may no longer be accessible. This work was supported in part by the National Natural Science Foundation of China under Grant 61901240, in part by the Natural Science Foundation of Shandong Province, China, under Grant ZR2019BF042, in part by the Department of Education of Shandong Province, and in part by the China Postdoctoral Science Foundation under Grant 2017M612204. (Corresponding author: Guojia Hou).}% <-this % stops a space
\thanks{Yuemei Li, Guojia Hou, and Zhenkuan Pan are with the College of Computer Science and Technology, Qingdao University, Qingdao 266071, China (email: lymqdu@126.com; guojiahou@qdu.edu.cn, zkpan@126.com).}% <-this % stops a space
% <-this % stops a space
\thanks{Peixian Zhuang is with the Key Laboratory of Knowledge Automation for Industrial Processes, Ministry of Education, the School of Automation and Electrical Engineering, University of Science and Technology Beijing, Beijing 100083, China (e-mail: zhuangpeixian0624@163.com).}% <-this % stops a space
}
%{Shell \MakeLowercase{\textit{et al.}}:  }

\maketitle
\begin{abstract}
Underwater images are typically characterized by color cast, haze, blurring, and uneven illumination due to the selective absorption and scattering when light propagates through the water, which limits their practical applications. Underwater image enhancement and restoration (UIER) is one crucial mode to improve the visual quality of underwater images. However, most existing UIER methods concentrate on enhancing contrast and dehazing, rarely pay attention to the local illumination differences within the image caused by illumination variations, thus introducing some undesirable artifacts and unnatural color. To address this issue, an effective variational framework is proposed based on an extended underwater image formation model (UIFM). Technically, dual high-order regularizations are successfully integrated into the variational model to acquire smoothed local ambient illuminance and structure-revealed reflectance in a unified manner. In our proposed framework, the weight factors-based color compensation is combined with the color balance to compensate for the attenuated color channels and remove the color cast. In particular, the local ambient illuminance with strong robustness is acquired by performing the local patch brightest pixel estimation and an improved gamma correction. Additionally, we design an iterative optimization algorithm relying on the alternating direction method of multipliers (ADMM) to accelerate the solution of the proposed variational model. Considerable experiments on three real-world underwater image datasets demonstrate that the proposed method outperforms several state-of-the-art methods with regard to visual quality and quantitative assessments. Moreover, the proposed method can also be extended to outdoor image dehazing, low-light image enhancement, and some high-level vision tasks. The code is available at https://github.com/Hou-Guojia/UDHTV. 
\end{abstract}

\IEEEPARstart{W}{ith} the continuous exploitation of marine resources, underwater vision plays a vital role in various marine applications and services, such as underwater target tracking, underwater autonomous exploration, and deep-sea debris detection. Underwater images have become the main carrier for perceiving and understanding underwater environments because they contain a wealth of valuable information including color, structure, and texture. However, unlike terrestrial images, underwater images captured in water often exhibit various degradations (i.e., color cast, haze, low contrast, and detail blurring) due to the complexity of underwater optical properties. Specifically, the selective absorption of light will affect the color rending \cite{01}. For example, since the light propagates through the water, red light with the longest wavelength is absorbed faster than the green and blue light, and thus the underwater images  often have a bluish or greenish appearance. The scattering also inevitably occurs when the suspended solids change the direction of light propagation, resulting in dim and blur effects. In addition, the widespread use of artificial light sources severely interferes with underwater imaging, which is easy to cause local illumination differences within the image. As a result, these degradations affect the extraction of valuable information and hinder further applications of engineering or research.

To improve the visibility of underwater images, a series of efforts have been made ranging from traditional methods (e.g., model-free \cite{02}, \cite{03}, \cite{04}, \cite{05} and model-based \cite{06}, 
 \cite{07}, \cite{08}, \cite{09}, \cite{10}, \cite{11},  to data-driven methods \cite{12}, \cite{13}, \cite{14}, \cite{15}, \cite{16}). Among them, the model-free methods aim to generate visually appealing outcomes by directly adjusting the distribution of pixel intensity without considering the physical characteristics of underwater imaging. Although these model-free methods can correct color and enhance contrast to some extent, the enhanced results tend to be over-enhanced and over-saturated. On the contrary, model-based methods depend on the underwater image formation model (UIFM) to estimate the relevant parameters, and a high-quality underwater image can be acquired by inverting the degradation process. However, since single assumption or water properties, these approaches cannot provide sufficient flexibility to effectively cope with the complex and changeable underwater environments. Accordingly, some attempts \cite{17}, \cite{18}, \cite{19} are starting to focus on extending the UIFM to ease the influence of the environment on image degradation. Despite the recovered images are not always satisfactory than expected, there is no doubt that extending physical imaging model is conducive for tackling the specific degradations of underwater images. In recent years, data-driven methods have shown remarkable performance in underwater image enhancement and restoration, benefiting from abundant data and powerful computational capabilities. Nevertheless, the requirements for large-scale training data, network structure design, and the generalization problem, may limit the practical implementation of these techniques. 

Fortunately, with the great success of variational technologies in natural image processing, including image reconstruction \cite{20}, image segmentation \cite{21} and image dehazing \cite{22}, it has also brought new strategies and perspectives to solve the quality degradation of underwater images. With regarding the Retinex theory-based and UIFM-based framework as the main genre, the variational methods specially designed for underwater images can not only restore the image with a pleasant visual appearance but also effectively enhance the structure and texture information of the image. However, existing variational methods \cite{23}, \cite{24}, \cite{25}, \cite{26} are mostly based on low-order regularization, in some cases, some undesired artifacts may be introduced in the progress of preserving the sharp structural edges. More recently, the corresponding high-order variational works \cite{27}, \cite{28}, \cite{29} are emerging to overcome this issue. Unfortunately, these variational models relying on Retinex theory or UIFM, only consider partial degradation factors. Therefore, it remains a challenge to comprehensively tackle the multiple degradations at the same time. 

For restoring underwater image by simultaneously correcting color, enhancing contrast, and removing haze, we propose a novel underwater image variation model based on an extended UIFM. The main contributions of our work are summarized as follows:

\begin{enumerate}[(a)]
    \item We establish an extended UIFM by introducing the local ambient illumination to overcome the limitations of local illumination differences. Based on this model, a novel variational framework with dual high-order regularizations is formulated for underwater images restoration, in which we design an Euler's elastica (EE) regularization and a Laplacian regularization to reveal the structure details in the reﬂectance and enforce the piece-wise smoothness on local ambient illumination, respectively. 
    \item A weight factors-based color compensation method is proposed to adaptively compensate the severely attenuated color channels of underwater images, which fully accounts for the pixel distribution of the attenuated color channels to avoid over-compensation and under-compensation. 
    \item In order to accurately estimate the local ambient illumination, an effective algorithm is developed by combining the local brightest pixel estimation and an improved gamma correction strategy.
    \item For solving the non-smooth optimization problem of the variational model,  an iterative optimization scheme based on the ADMM is exploited  to expedite the whole process. 
\end{enumerate}

The rest of this paper is structured as follows. Section
\ref{Related worlks}  summarizes the related work about model-based methods. Section
\ref{Proposed methods} presents a detailed description of the proposed method. The experimental evaluations and discussions and potential applications are reported in Section 
\ref{Experimental results and discussion}. Finally, Section 
\ref{CONCLUSION} concludes the paper.

\section{Related worlks}\label{Related worlks}
Since the proposed variational method is based on the extended underwater image formation model, following, we provide a detailed overview of the related work associated with model-based methods. These methods can generally be divided into the two categories: imaging model-based methods and extended imaging model-based methods.
\begin{figure*}[!t]
\centering
\includegraphics[width=\linewidth]{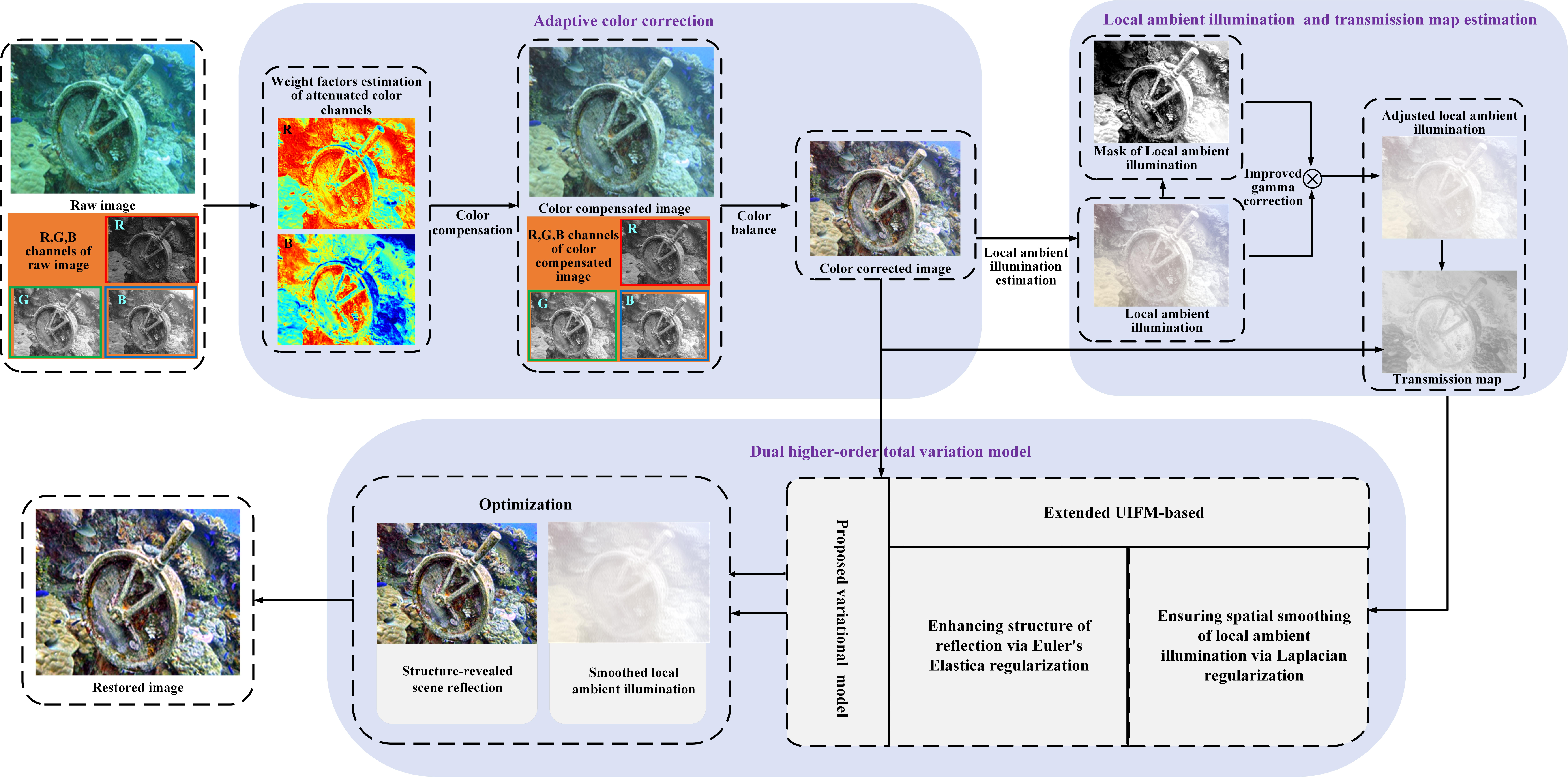}
\caption{Flowchart of the proposed method.}
\label{Fig1}
\end{figure*}
\subsection{Imaging Model-based Methods}
Due to the similarity between UIFM and the foggy image formation model, various prior knowledge-guided methods derived from dark channel priors (DCP) \cite{30} have been developed to recover haze-free images. Considering that red light decays fastest in underwater media, Galdran et al. \cite{31} presented a red channel prior (RCP) method, which utilized the inverted red channel and green-blue channels to calculate the dark channel. Subsequently, Liang et al. \cite{32} developed a generalized underwater dark channel prior (GUDCP)
method, in which multiple priors were incorporated to estimate
the transmission map. However, the DCP-based methods are susceptible to light attenuation and artificial light sources in underwater scenes. Therefore, some researchers focus on reasoning scene depth information to obtain a more robust estimation of transmission map. For example, Peng et al. \cite{33} estimated the scene depth and background light using image blurriness and light absorption (IBLA). Afterward, Berman et al. \cite{09} considered the difference in attenuation coefficients of the direct component and backward scattering component, and the transmission map was obtained via backscatter pixel prior.

More recently, model-based deep learning methods have been widely applied in underwater image processing, which has inherited the advantages of physical imaging model and deep learning strategies. In \cite{34}, Liu et al. introduced an integrated learning framework termed IPMGAN that embedded the revised physical model into the network architecture for underwater image enhancement. To further exploit the domain knowledge of underwater imaging, Li et al. \cite{35} developed a multi-color space encoder network to extract the discriminative features and designed a transmission-guided decoder network to emphasize the quality-degraded regions. Afterward, Hambarde et al. \cite{36} presented a novel cascaded-stream network in which both the coarse-level and fine-level depth maps were estimated from a single underwater image. Subsequently, Song et al. \cite{37} proposed a dual-model methodology consisting of two branches: the revised imaging network model and the visual perception correction model, to estimate the underwater scene parameters and correct the degraded color. Lately, Cong et al. \cite{38} established a physical model-guided generative adversarial network (PUGAN) to reconstruct a high-quality underwater image by employing two subnetworks and a dual discriminator. However, despite the above strategies that can significantly improve the robustness and generalization of the deep enhancement model, there still exists a shortage of real-world paired training data to effectively train deep networks applicable to various degradation types. 

Also, several studies are currently focusing on variational strategies to improve the perceptual quality of underwater image. Combining the improved UDCP and UIFM, Hou et al. \cite{24} introduced a total variation regularization into the underwater image variational model to achieve good performance in image dehazing and denoising. Later, Ding et al. \cite{25} designed a depth-aware based variation model, which fully considered the characteristics of the scene depth map to reduce haze and enhance structure. Subsequently, Li et al. \cite{26} presented a new variational framework for underwater image restoration, in which they estimated the transmission map using the intrinsic boundary constraint and combined the local variance of the exponential mean with the external prior to preserve image edge and remove noise.  More recently, Hao et al. \cite{29} developed a variational model with high-order regularization to achieve image dehazing. They constructed a brightness-aware approach that fused multiple priors to estimate the transmission map and utilized high-order regularization to restore fine-scale details. Although certain breakthroughs have been made, the above methods fail to comprehensively analyze the factors that affect the imaging quality.
\subsection{Extended Imaging Model-based Methods}
Unlike the predominant approaches that primarily consider the direct component and backward scattering component, some researchers have additionally incorporated the forward scattering component into the UIFM. Cheng et al. \cite{39} proposed a hybrid method for underwater image dehazing and deblurring, in which a red-dark channel prior was defined to estimate the transmission, and a low-pass filter was developed to eliminate the fuzziness in underwater images via analyzing the physical properties of the point diffusion function. However, this method fails to provide high-quality results and tends to introduce some visually displeasing red artifacts. To overcome this issue, Park et al. \cite{17} utilized the inverse relationship between the scene distance and the geodesic color distance from the background light to estimate the transmission map. In addition, they also approximated the point spread function based on the complete UIFM, achieving good performance in visual quality improvement and artifact reduction. In the current work, Xie et al.\cite{18} designed a novel variation framework for underwater image restoration, in which they also integrated the complete UIFM into a normalized total variation model to obtain blur kernel and employed RCP and quad-tree subdivision algorithm to estimate the transmission map and background light. 

Accordingly, another kind of extended model-based methods have also made great progress with considering the light propagation process along both the source-scene path and the scene-sensor path. For instance, Ding et al. \cite{19} developed a two-step progressive-based restoration method, in which they decomposed the complex restoration process into two subproblems and formulated the distinct energy functions guided by prior knowledge to remove the light attenuation along dual-path. Differently, based on a novel image formation model called GIFM, Liang et al. \cite{10} first established an objective energy function to simultaneously acquire color distorted component and color corrected component in a unified manner, and then designed an adaptive weighted fusion strategy to eliminate the haze in the color corrected component. Despite these methods can solve multiple types of degradation problems simultaneously, they are not always effective in some challenging scenarios, such as severe color cast, high backscatter, and complex lighting.

\section{Proposed method}\label{Proposed methods}
In this section, we first present an extended UIFM to break through the limitations of local illumination differences. Then, based on the extended UIFM, we develop a novel underwater variation model via double high-order regularizations. In our framework, a two-step color correction strategy is first presented that combines adaptive color compensation and color balance to remove the color cast. Also, the local ambient illuminance and transmittance are estimated by employing the local patch brightest pixel and performing min-operator on the extended UIFM, respectively. Moreover, an optimal algorithm based on ADMM is designed to solve the optimization problem of our proposed variational model. The flowchart of the proposed method is presented in Fig. \ref{Fig1}.

\subsection{Model Construction}
Generally, most model-based methods usually represent the degraded underwater image based on the simplified UIFM \cite{40,41}, which describes the propagation process of light. According to the simplified UIFM, the light detected by the camera is mainly composed of two components. One is the light that is reflected from the surface of the object received by the camera (direct component). The other is the light that is scattered by the non-target object (backward scattering). Thus, the simplified UIFM can be expressed as:
\begin{equation}
\label{equation1}
\setlength{\abovedisplayskip}{3pt}
I(x)=\underbrace{J(x) t(x)}_{\text {Direct Component }}+\underbrace{B(x)(1-t(x))}_{\text {Backward Scattering }} , 
\setlength{\belowdisplayskip}{3pt}
\end{equation}
where $ I(x) $ represents the captured image, $ J(x) $ refers to the undistorted  image, $ t(x) $ refers to the medium transmission maps, $ B(x) $ is the global background light.

Actually, in some challenging real underwater scenes, the widely-used global background light cannot capture the difference of local illumination in the image, which may result in undesirable outcomes (e.g., over/under-enhancement, uneven brightness, and haze residue), especially for the scenes with complex illumination. Here, instead of using global background light, local ambient illumination $ L(x) $ is introduced into (\ref{equation1}). In addition, given that the UIFM is highly similar to the outdoor fogging model. Inspired by \cite{42}, the UIFM can be extended as:
\begin{equation}
\label{equation2}
\setlength{\abovedisplayskip}{3pt}
I(x)=R(x)L(x)t(x)+L(x)(1-t(x)).
\setlength{\belowdisplayskip}{3pt}
\end{equation}

Depending on the extended UIFM, we propose a novel underwater high-order total variation model consisting of Euler’s elastica regularization and Laplacian regularization to simultaneously recover the scene reflection $ R(x) $ and the local ambient illumination $ L(x) $. The proposed energy function is formulated as:
\begin{equation}
\begin{aligned}
\label{equation3}
\begin{array}{l}
E(R, L)=\frac{1}{2} \int_{\Omega}(R L \mathrm{t}+L(1-t)-I)^{2} d x \\
+\int_{\Omega}\left(\alpha+\beta\left(\nabla \cdot\left(\frac{\nabla R}{|\nabla R|}\right)\right)^{2}\right)|\nabla R| d x+\frac{\gamma}{2} \int_{\Omega}|\Delta L|^{2} d x
\end{array},
\end{aligned}
\end{equation}
where $ \alpha $, $ \beta $, and $ \gamma $ are the positive parameters that control the relationships of regularization terms. $ \nabla $ and $ \Delta $ denote the first-order and second-order differential operators, respectively. The first term  $ (RLt+L(1-t)-I)^{2} $ is the data item, providing the fidelity between the original underwater image $ I $ and the recomposed one $ RLt+L(1-t) $. The second term $ \left(\alpha+\beta\left(\nabla \cdot\left(\frac{\nabla R}{|\nabla R|}\right)\right)^{2}\right)|\nabla R| $ corresponds to the Euler's elastica regularization, which encourages edge-preserving and piecewise continuous of the reflectance component  $ R $. The last term $ |\Delta L|^{2} $ pertains to the Laplacian regularization. Specifically, the ideal ambient illumination layer aims for a piecewise smooth quality while preserving prominent edges. To achieve this, second-order Laplacian regularization is employed to approximate the piecewise linear composition, thereby preserving only the primary edges of the ambient illumination. Simultaneously, the utilization of the $L_{2}$ norm ensures sparsity in the second-order gradient of the ambient illumination, promoting piecewise smoothness in the illumination layer.

\begin{figure}[!t]
\setlength{\abovecaptionskip}{-0.1cm}
\setlength{\belowcaptionskip}{-1cm}
\centering
\includegraphics[width=\linewidth]{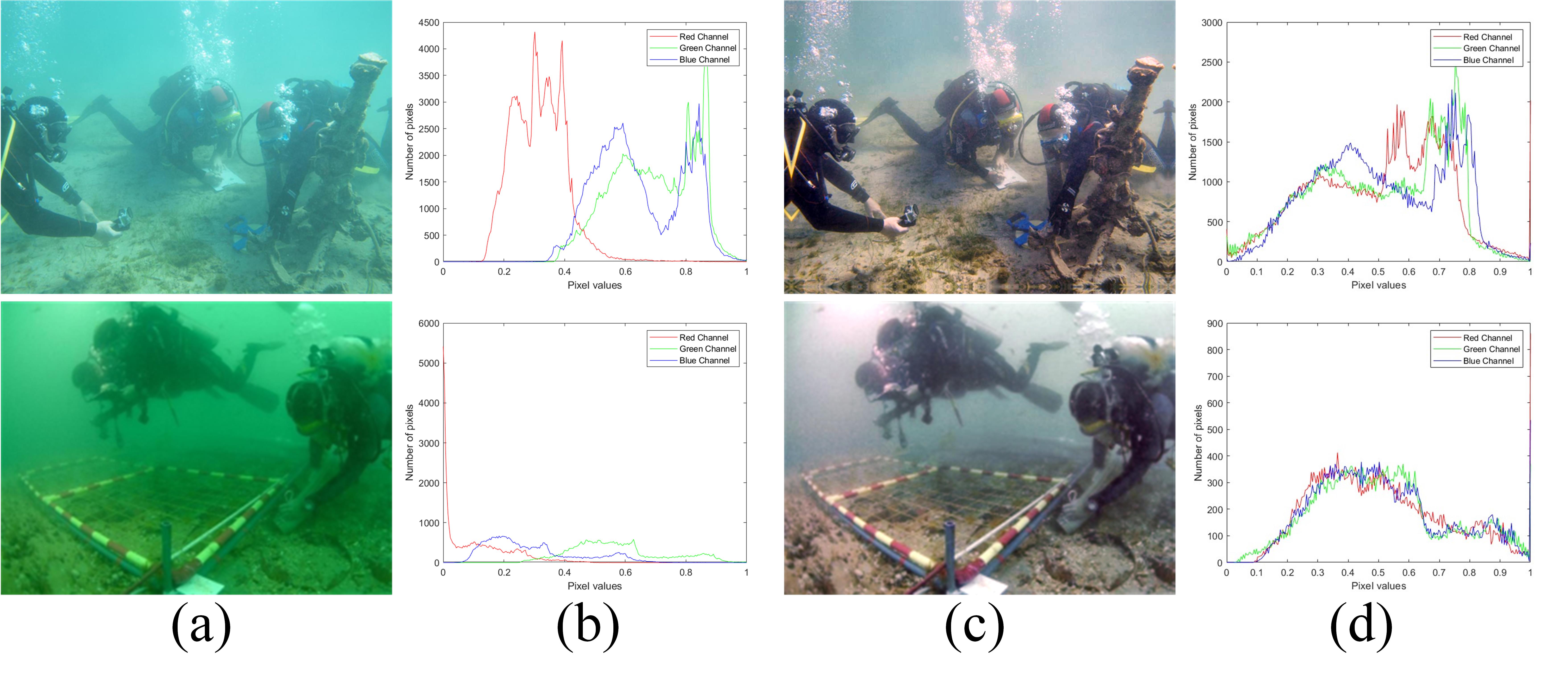}
\caption{Examples of color correction. (a) Raw images, (b) the corresponding three-color histogram distributions of (a), (c) the enhanced results using the proposed color correction method, (d) the corresponding three-color histogram distributions of (c).}
\vspace{-3mm}
\label{Fig2}
\end{figure}

\begin{figure}[!t]
\setlength{\abovecaptionskip}{0cm}
\setlength{\belowcaptionskip}{-1cm}
\centering
\includegraphics[width=\linewidth]{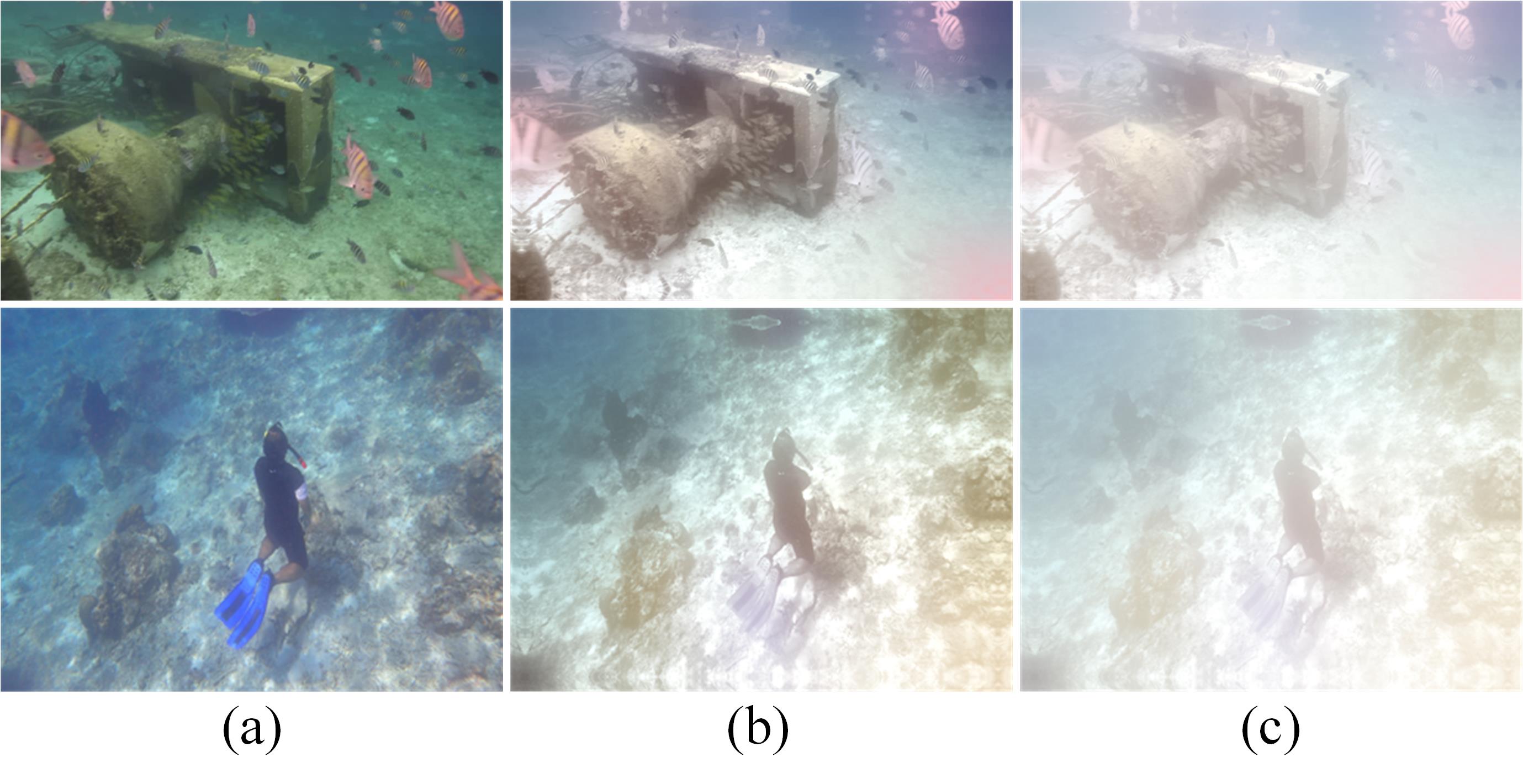}
\caption{Examples of estimating the local ambient illumination $ L $. (a) Raw images, (b) the local ambient illumination of $ L $, (c) the refined local ambient illumination $ L $ of (b). }
\vspace{-3mm}
\label{Fig3}
\end{figure}
\subsection{Adaptive Color Correction}
In reality, the images still suffer from color cast due to the selective attenuation of light as it propagates in the underwater medium. Strong attenuation not only occurs in the red channel with the longest wavelength but also sometimes appears in the other color channels. For example, in turbid water scene or the scene with a high concentration of plankton, the attenuation coefficient of the green channel is smaller than that of the other two (red and blue) channels. On the contrary, for the deep-water scene, the attenuation coefficient of the blue channel is smaller. Based on this observation, we first propose an adaptive color compensation algorithm by performing different operations to compensate the other two severely attenuated color channels via the average value of the blue or green channels. In one case ($ \bar{I}_{g}>\bar{I}_{b} $), the green channel is used to compensate for the red and blue channels, as follows:
\begin{equation}
\begin{aligned}
\label{equation4}
\setlength{\abovedisplayskip}{-3pt}
\setlength{\belowdisplayskip}{-3pt}
\left\{\begin{array}{l}I_{r c}(x)=I_{r}(x)+d * \omega_{r}\left(\bar{I}_{g}-\bar{I}_{r}\right) * I_{r}(x) 
\\I_{b c}(x)=I_{b}(x)+d * \omega_{b}\left(\bar{I}_{g}-\bar{I}_{b}\right) * I_{b}(x) \end{array},\right.
\end{aligned}
\end{equation}

where $ I_{r}, I_{g}, I_{b} $  represent the red, green, and blue channels of the image $ I $, respectively. Meanwhile, $ \omega_{r}=\left(1- \newline \operatorname{sigmod}\left(I_{r}\right)\right)^{2} $, $ \omega_{b}=\left(1-\operatorname{sigmod}\left(I_{b}\right)\right)^{2} $ are the weight factors that adaptively decrease (increase) as pixel values become larger (smaller), thus preventing over-compensation and under-compensation of red channel and blue channel. $ d $ is an empirical constant, set to 5. Accordingly, in another case ($ \bar{I}_{b}>\bar{I}_{g} $), the blue channel is used to compensate for the red and green channels, as follows:
\begin{equation}
\begin{aligned}
\label{equation5}
\left\{\begin{array}{l}I_{r c}(x)=I_{r}(x)+d * \omega_{r}\left(\bar{I}_{b}-\bar{I}_{r}\right) * I_{r}(x) \\I_{g c}(x)=I_{g}(x)+d^{*} \omega_{g}\left(\bar{I}_{b}-\bar{I}_{g}\right) * I_{g}(x)\end{array},\right. 
\end{aligned}
\end{equation}
where $ \omega_{g}=\left(1-\operatorname{sigmod}\left(I_{g}\right)\right)^{2} $ is the weight factor to prevent over-compensation and under-compensation of the green channel. After compensating for the attenuation channels, we further perform a simple yet effective operation \cite{23} to obtain the color balance image:
\begin{equation}
\begin{aligned}
\label{equation6}
I_{c b}^{\mathrm{c}}=\frac{255}{2}\left(1+\frac{I^{\mathrm{c}}-I_{\text {mean }}^{\mathrm{c}}}{\varphi I_{\mathrm{var}}^{\mathrm{c}}}\right),
\end{aligned}
\end{equation}
where $ I_{\text {mean }}^{\mathrm{c}} $, $  I_{\text {var }}^{\mathrm{c}} $ are the mean value and variance in each channel. $  I_{c b}^{\mathrm{c}} $ is the enhanced image after color balance, $ \varphi $ is empirical constant, set to 2.3. Fig. \ref{Fig2} shows two degraded underwater images and their enhanced results using our adaptive color correction method, along with the associated histogram distributions. It can be seen that our color correction method not only can effectively reduce the color cast but also broaden the distribution of the three-color histograms broaden.
\vspace{-3mm}

\subsection{ Local Ambient Illumination and Transmission Estimation}
To solve (\ref{equation3}), we first need to estimate the local ambient illumination $ L $ and the transmission $ t $. In our framework, the initial value of local ambient illumination is picked by the brightest pixel in the local block size (5 × 5). Then, inspired by the mask $ M_{L} $ that perceives the brightness information \cite{43}, we employ an improved gamma correction strategy to adaptively adjust the distribution of $ L $. Formally, the expression of $ M_{L} $ is:
\begin{equation}
\begin{aligned}
\label{equation7}
M_{L}=\left\{\begin{array}{cc}
1 & \text { mean }\left(L^{\mathrm{c}}(x, \mathrm{y})>\operatorname{mean}(L)\right) \\
0 & \text { otherswise }
\end{array} .\right.
\end{aligned}
\end{equation}

To avoid artifacts, the $ M_{L} $ is further refined by the guided filter \cite{44}, and the adjusted $ M_{L} $ can be obtained by:
\begin{equation}
\begin{aligned}
\label{equation8}
\setlength{\abovedisplayskip}{-3pt}
\setlength{\belowdisplayskip}{-3pt}
\text { gamma }=1-\delta * \theta^{\frac{M_{L}-\operatorname{mean}\left(M_{L}\right)}{\max \left(M_{L}\right)-\min \left(M_{L}\right)}} \text {, }
\end{aligned}
\end{equation}
\begin{equation}
\begin{aligned}
\label{equation9}
L=L^\text { gamma },
\end{aligned}
\end{equation}
where $ \theta $, and $ \delta $ are two constants, set to 0.8 and 0.5, respectively. Here, we also give two examples to demonstrate the effectiveness of our strategy, as shown in Fig. \ref{Fig3}. We can observe that the illumination of Fig. \ref{Fig3}(c) is more uniform than Fig. \ref{Fig3}(b). 
After obtaining the local ambient illumination $ L $ , we further perform a min-operator on both sides of (\ref{equation2}) to estimate the transmission $ t $ based on He’s method \cite{30} , and the formula of the transmission is expressed as:
\begin{equation}
\begin{aligned}
\label{equation10}
t^{c}(x) & = 1-\min _{y \in \Omega(x)}\left\{\min _{c \in R, G, B} \frac{I^{c}(y)}{L^{c}(x)}\right\},
\end{aligned}
\end{equation}
where $\Omega(x)$ is a local patch centered at x. Finally, we refine it using the guided filter.

\subsection{ Numerical Solution}
For solving the ill-posed problem of (\ref{equation3}), we design a fast numerical algorithm based on the alternating direction multiplier method (ADMM) \cite{45} to accelerate the progress of the optimal solution. Table \ref{table1} presents an overview of commonly used symbols. By adding six auxiliary vector variables: $ \vec{w}=\nabla R $, $ \vec{p}=\frac{\vec{w}}{|\vec{w}|} $, $ \vec{p}=\vec{q} $, $ v=\nabla \cdot \vec{p} $, $ \vec{m}=\nabla L $ and $ g=\nabla \cdot \vec{m} $, (\ref{equation3}) can be translated into:
\begin{equation}
\begin{aligned}
\label{equation11}
\begin{array}{l}E(R, L, \vec{w}, v, \vec{p}, \vec{q}, \vec{m}, g)= \\\left\{\begin{array}{l}\frac{1}{2} \int_{\Omega}((R L \mathrm{t}+L(1-t)-I))^{2} d x+\int_{\Omega}\left(\alpha+\beta v^{2}\right)|\vec{w}| d x \\+\frac{\gamma}{2} \int_{\Omega}|g|^{2} d x+\frac{\mu_{1}}{2} \int_{\Omega}\left|\vec{w}-\nabla R+\frac{\vec{\lambda}_{1}}{\mu_{1}}\right|^{2} d x \\+\int_{\Omega}\left(\lambda_{2}+\mu_{2}\right)(|\vec{w}|-\vec{w} \cdot \vec{q}) d x+\frac{\mu_{3}}{2} \int_{\Omega}\left|v-\nabla \cdot \vec{p}+\frac{\lambda_{3}}{\mu_{3}}\right|^{2} d x \\+\int_{\Omega} \frac{\mu_{4}}{2} \int_{\Omega}\left|\vec{p}-\vec{q}+\frac{\vec{\lambda}_{4}}{\mu_{4}}\right|^{2} d x + \frac{\mu_{5}}{2} \int_{\Omega}\left|\vec{m}-\nabla L+\frac{\vec{\lambda}_{5}}{\mu_{5}}\right|^{2} d x \\+\frac{\mu_{6}}{2} \int_{\Omega}\left|g-\nabla \vec{m}+\frac{\lambda_{6}}{\mu_{6}}\right|^{2} d x\end{array}\right\} .\end{array}
\end{aligned}
\end{equation}

 To efficiently address the above constrained optimization problem, we decompose the problem (\ref{equation11}) into eight sub-problems. Then, the equivalent energy function can be solved by updating one variable while fixing the other unrelated variable, as
follows:\\
 \noindent\textbf{(P1):} Subproblems of $R$ and $L$ reconstruction:
\begin{subequations}
\begin{align}
\varepsilon_{1}(R) &= \min _{R}\left\{\begin{array}{l}
\frac{1}{2} \int_{\Omega}((R L \mathrm{t}+L(1-t)-I))^{2} d x \\
+\frac{\mu_{1}}{2} \int_{\Omega}\left|\vec{w}-\nabla R+\frac{\vec{\lambda}_{1}}{\mu_{1}}\right|^{2} d x
\end{array}\right\},
\tag{12-a}
\label{12-a} \\
\varepsilon_{2}(L) &= \min _{L}\left\{\begin{array}{l}
\frac{1}{2} \int_{\Omega}((R L \mathrm{t}+L(1-t)-I))^{2} d x \\
+\frac{\mu_{5}}{2} \int_{\Omega}\left|\vec{m}-\nabla L+\frac{\vec{\lambda}_{5}}{\mu_{5}}\right|^{2} d x
\end{array}\right\}.
\tag{12-b}
\label{12-b}
\end{align}
\end{subequations}
 \noindent\textbf{(P2):} Subproblems involving auxiliary variables for $R$ reconstruction:
\begin{subequations}
\begin{align}
\varepsilon_{3}(\vec{w}) &= \underset{\vec{w}}{\arg \min }
\left\{
\begin{array}{l}
\int_{\Omega}\left(\alpha+\beta v^{2}\right)|\vec{w}| d x \\
+\frac{\mu_{1}}{2} \int_{\Omega}\left|\vec{w}-\nabla R+\frac{\vec{\lambda}_{1}}{\mu_{1}}\right|^{2} d x \\
+\int_{\Omega}\left(\lambda_{2}+\mu_{2}\right)(|\vec{w}|-\vec{w} \cdot \vec{q}) d x
\end{array}
\right\},
\tag{13-a}
\label{13-a}\\
\varepsilon_{4}(\vec{p}) &= \underset{\vec{p}}{\arg \min }
\left\{
\begin{array}{l}
\frac{\mu_{3}}{2} \int_{\Omega}\left|v-\nabla \cdot \vec{p}+\frac{\lambda_{3}}{\mu_{3}}\right|^{2} d x \\
+\frac{\mu_{4}}{2} \int_{\Omega}\left|\vec{p}-\vec{q}+\frac{\vec{\lambda}_{4}}{\mu_{4}}\right|^{2} d x
\end{array}
\right\},
\tag{13-b}
\label{13-b}
\end{align}
\end{subequations}
\begin{subequations}
\begin{align}
\varepsilon_{5}(\vec{q}) &= \underset{\vec{q}}{\arg \min }
\left\{
\begin{array}{l}
\int_{\Omega}\left(\lambda_{2}+\mu_{2}\right)(|\vec{w}|-\vec{w} \cdot \vec{q}) d x \\
+\frac{\mu_{4}}{2} \int_{\Omega}\left|\vec{p}-\vec{q}-\frac{\vec{\lambda}_{4}}{\mu_{4}}\right|^{2} d x
\end{array}
\right\},
\tag{13-c}
\label{13-c}\\
\varepsilon_{6}(v) &= \underset{v}{\arg \min }\left\{
\begin{array}{l}
\int_{\Omega}\left(\alpha+\beta v^{2}\right)|\vec{w}| d x+ \\
\frac{\mu_{3}}{2} \int_{\Omega}\left|v-\nabla \cdot \vec{p}+\frac{\lambda_{3}}{\mu_{3}}\right|^{2} d x
\end{array}
\right\}.
\tag{13-d}
\label{13-d}
\end{align}
\end{subequations}

\noindent\textbf{(P3):} Subproblems involving auxiliary variables for $L$ reconstruction:
\begin{subequations}
\begin{align}
\varepsilon_{7}(\vec{m}) &= \underset{\vec{m}}{\arg \min }\left\{
\begin{aligned}
&\frac{\mu_{5}}{2} \int_{\Omega}\left|\vec{m}-\nabla L+\frac{\vec{\lambda}_{5}}{\mu_{5}}\right|^{2} d x+ \\
&\frac{\mu_{6}}{2} \int_{\Omega}\left|g-\nabla \vec{m}+\frac{\lambda_{6}}{\mu_{6}}\right|^{2} d x
\end{aligned}
\right\},
\tag{14-a}
\label{14-a}
\end{align}
\end{subequations}
\begin{align}
\varepsilon_{8}(g) = \underset{g}{\arg \min }\left\{\frac{\gamma}{2} \int_{\Omega}|g|^{2} d x+\frac{\mu_{6}}{2} \int_{\Omega}\left|g-\nabla \cdot \vec{m}+\frac{\lambda_{6}}{\mu_{6}}\right|^{2} d x\right\}.
\tag{14-b}
\label{14-b}
\end{align}
\begin{table}[!t]
\tiny
\caption{The significance of pivotal symbols}
\label{table1}
\centering
\resizebox{\linewidth}{!}{ 
\scalebox{1}{
\begin{tabular}{ll} 
\hline
Symbols & Explanation \\ \hline
$\Omega, \partial \Omega$       & The entire domain and boundary of the image, respectively           \\
$\nabla, \Delta$       & The first-order and second-order differential operators, respectively           \\
$\|\cdot\|$       & The absolute value           \\
$\alpha, \beta, \gamma$       & The positive penalty parameters of regularization terms           \\
$\vec{w}, \vec{p}, \vec{q}, v, \vec{m}, g$       & The auxiliary variables of augmented Lagrangian function          \\
$\{\mu\}_{i=1}^{6}$      & The positive penalty parameters of augmented Lagrangian function          \\
$\vec{\lambda}_{1}, \lambda_{2}, \lambda_{3}, \vec{\lambda}_{1}, \vec{\lambda}_{1}, \lambda_{6}$      & The Lagrange multipliers of augmented Lagrangian function          \\
$FFT, FFT^{-1}$      & The fast Fourier transform and its inverse transformation, respectively          \\ \hline
\end{tabular}}}
\end{table}

The mathematical solutions corresponding to (P1), (P2), and (P3) are provided in the \hyperref[Appendix]{Appendix}. 

\section{Experimental results and discussion}\label{Experimental results and discussion}
In this section, we first evaluate the performance of our proposed method by performing qualitative and quantitative comparisons with several state-of-the-art (SOTA) methods. Then, some ablation studies are conducted to demonstrate the contribution of each key component of the proposed method. Finally, we extend the proposed method to other potential applications. In the following experiments, the  parameters $\alpha$, $\beta$, $\gamma$  are empirically set to $1e-3$, $1e-3$ and 10. 

\textbf{Comparison Methods}. We compare our method against eleven SOTA methods, including RCP \cite{31}, IBLA \cite{33}, low-light underwater image enhancer ($ \text{L}^{\text{2}}\text{UWE} $) \cite{05}, statistical model of background light and optimization of transmission (SMBLOT) \cite{08}, underwater convolutional neural network (UWCNN) \cite{46}, Haze-Lines \cite{09} Bayesian retinex (BR) \cite{27}, minimal color loss and locally adaptive contrast enhancement (MLLE) \cite{02}, hyper-Laplacian reflectance priors (HLRP) \cite{28}, underwater Laplacian variation (ULV) \cite{29} and contrastive semi-supervised learning for underwater image restoration (Semi-UIR)\cite{47}. Among them, RCP, IBLA, SMBLOT, Haze-Lines, and ULV are model-based methods, while $ \text{L}^{\text{2}}\text{UWE} $, BR, MLLE, and HLRP are model-free methods. UWCNN and Semi-UIR are data-driven methods.

\textbf{Benchmark Datasets}. Our proposed method is tested on three underwater image datasets (UIEB \cite{48}, UIQS \cite{49}, and Color-Checker7 \cite{50}. The UIEB dataset contains 890 degraded images captured under multiple challenging scenes, including greenish/bluish scene, low-light scene, turbid and hazy scene, and deep-water scene. Unlike the UIEB dataset, which is collected from the Internet and self-captured videos, the entire UIQS dataset is captured using water-proof video cameras in Zhangzi Island. It is further divided into five subsets based on the image quality evaluation index. The Color-Checker7 dataset contains 7 underwater images that are taken by different cameras in a shallow swimming pool. Each image in this dataset also includes a Color checker.

\textbf{Evaluation Metrics}. To objectively evaluate the performance of the proposed method, we employ several widely used image quality assessment (IQA) metrics including fog aware density evaluator (FADE) \cite{51}, Entropy, underwater color image quality evaluation (UCIQE) \cite{52}, frequency domain underwater metric (FDUM) \cite{53}, and CIEDE2000 \cite{54}. FADE relies on measurable deviations from statistical regularities to predict perceptual fog density. Entropy measures the average information contained in an image, which expresses the richness of detail. UCIQE and FDUM are specifically designed for evaluating underwater image quality. UCIQE is generated by linearly combing chroma, saturation, and contrast to quantify image quality. FDUM is a more comprehensive measurement indicator, which quantifies three aspects of underwater images including colorfulness, contrast, and sharpness. The higher values of Entropy, UCIQE, and FDUM represent better results. Conversely, a lower FADE represents a better outcome. Different from other IQA metrics that evaluate values using global statistical features, CIEDE2000 only considers color difference based on the CIELab color space. Lower CIEDE2000 values indicate a smaller difference between the reference ground truth color and restored color.
\subsection{Color Correction Evaluation}
\begin{figure}[!t]
\setlength{\abovecaptionskip}{-0.2cm}
\setlength{\belowcaptionskip}{-0.2cm}
\centering
\includegraphics[width=\linewidth]{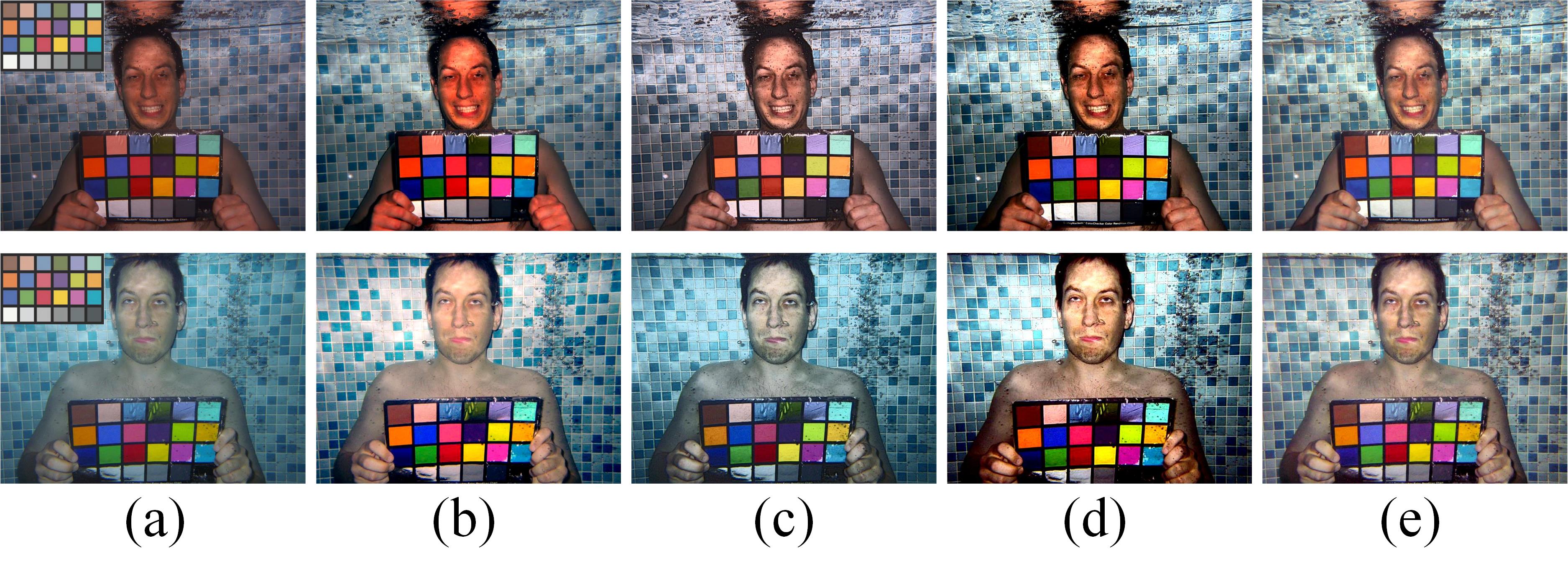}
\caption{Comparison of color correction on Color-Checker7. (a) Raw images, and the restored results of (b) SMBLOT \cite{08}, (c) BR \cite{27}, (d) HLRP \cite{28}, and (e) the proposed adaptive color correction method, respectively.}
\label{Fig4}
\end{figure}

To evaluate the performance of our proposed method for color correction, we conduct some visual comparisons on the Color-Checker7 dataset. For these images, we compare our adaptive color correction method against the other three color-correction based methods, their restored results are presented in Fig. \ref{Fig4}. It can be seen that SMBLOT fails to recover authentic color, and the resulting Color checker has an apparent reddish color deviation. Similarly, BR has less effect on color correction. Although HLRP has a certain recovery effect on color correction, the orange located at (2, 1) in the top image and the green located at (1, 4) in the bottom image of the Color checkers are distorted, leading to misleading results. On the contrary, our adaptive color correction method achieves remarkable performance in restoring realistic colors.  
In the qualitative evaluation, the measured values of CIEDE2000 for each image are reported in Table \ref{table2}. It is evident that our proposed method has the lowest CIEDE2000 scores on six cameras, and gains highly competitive scores on the remaining one camera, ultimately resulting in the best average score. All in all, for the underwater images captured by different cameras, our method has a strong robustness to eliminate color cast.
\begin{table*}[!t]
\small
\caption{Quantitative comparisons of CIEDE 2000 on the Color Checker 7. (The top two performers are highlighted in \textcolor{red}{red} and \textcolor{blue}{blue}, respectively)}
\label{table2}
\centering
\begin{tabular}{cccccccccc}
\hline
Methods & OlyT6000  & OlyT8000 & Can D10 & Pan TS1  & Pen W60  & FujiZ33    & Pen W80   & Average \\ \hline
SMBLOT  & \textcolor{blue}{13.569}  & \textcolor{blue}{13.587}   & 15.553 & 14.122  & 15.357  &  \textcolor{red}{12.140} & \textcolor{blue}{14.567}  & \textcolor{blue}{14.127}  \\
BR  & 16.032 & 16.605 & \textcolor{blue}{15.048}  &  \textcolor{blue}{12.340}   & \textcolor{blue}{14.005}  & 23.728  & 14.689   & 16.063  \\
HLRP  & 20.803  & 20.427  & 16.617  & 12.932  & 14.776  & 36.186  & 16.996  & 19.819   \\
Proposed  & \textcolor{red}{10.454} & \textcolor{red}{10.633} & \textcolor {red}{12.477} & \textcolor{red}{9.412} & \textcolor{red}{11.664} & \textcolor{blue}{14.333}  & \textcolor{red}{11.707} & \textcolor{red}{11.525}  \\ \hline
\end{tabular}
\end{table*}

\begin{figure*}[!t]
\centering
\setlength{\abovecaptionskip}{-0.2cm}
\includegraphics[width=\linewidth]{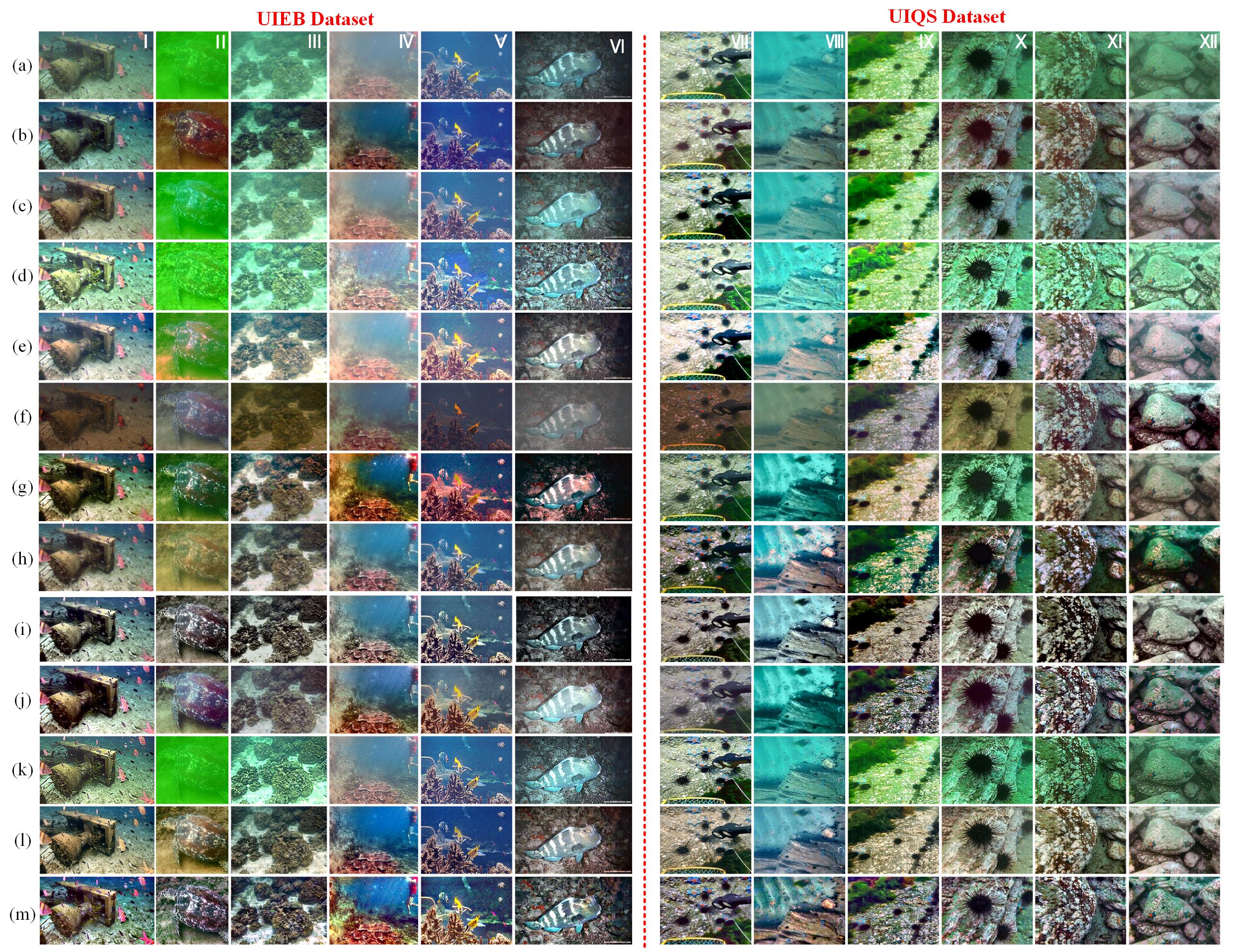}
\caption{Qualitative comparisons on various challenging scenes from the UIEB and UIQS datasets. (a) Raw images, the restored results of (b) RCP \cite{31}, (c) IBLA \cite{33}, (d) $ \text{L}^{\text{2}}\text{UWE} $ \cite{05}, (e) SMBLOT \cite{08}, (f) UWCNN \cite{46}, (g) Haze-Lines \cite{09}, (h) BR \cite{27}, (i) MLLE \cite{02}, (j) HLRP \cite{28}, (k) ULV \cite{29}, (l) Semi-UIR \cite{47}, and (m) the proposed method, respectively.}
\label{Fig5}
\end{figure*}
\subsection{Comprehensive Visual Quality Evaluation}
In this part, we first compare the enhanced results of different UIER methods on the UIEB. Due to the limited space, we select several representative underwater images with different degradations (i.e., hazy, greenish, turbid, low-visibility, bluish, and low-light images) from UIEB as examples. Their corresponding enhanced and restored results are presented in the left side of Fig. \ref{Fig5}. We can observe that RCP performs well in dehazing. However, it aggravates color deviation in greenish scene. Similarly, IBLA has less effects on color correction and contrast enhancement because the prior is invalidated on seriously distorted images. Although the $ \text{L}^{\text{2}}\text{UWE} $ and SMBLOT methods are effective in improving the brightness of images, both of them perform poorly in removing haze-like effects. In addition, $ \text{L}^{\text{2}}\text{UWE} $ introduces some undesirable artifacts and exhibits unnatural appearance. SMBLOT often leads to overexposure of some bright regions, especially for the underwater images captured in yellowish and turbid scenes. Likewise, UWCNN introduces some red artifacts. Furthermore, although Haze-Lines can enhance the image contrast to some extent, it struggles to completely remove the haze-like appearance. BR effectively eliminates color interference, while it is ineffective in turbid and low-light scenes. Likewise, although MLLE can significantly improve the visibility of degraded images, it falls short in performing well in high backscatter and low-light scenes. HLRP can correct blue-green scenes well, but its ability for structure improvement is limited. In contrast, ULV can enhance the image structure well. Unfortunately, it fails to produce a satisfactory result in complex ambient illumination scenes. Semi-UIR can effectively remove backscatter effects, but it produces unrealistic color restoration results in the blue-green scenes. Fortunately, our proposed method can simultaneously remove color deviation, reveal more structural information, as well as sharpen edge contour. Moreover, unlike other UIER methods, our approach can also better overcome the complex illumination problem benefiting from its consideration of local ambient illumination.
Following, we also compare the enhanced results of UIER methods on the UIQS dataset. The visual comparisons are shown  in the
right side of Fig. \ref{Fig5}. We can see that RCP, IBLA, Haze-Lines, and BR methods cannot generate good results since they are unsatisfactory for removing the color deviations. Moreover, some comparative methods introduce inappropriate artifacts, while also over-enhancing details in certain regions (e.g., $ \text{L}^{\text{2}}\text{UWE} $, UWCNN, and SMBLOT). Although HLRP and Semi-UIR are satisfactory for removing the color deviations, they still exhibit other quality defects. ULV and MLLE perform better than the above methods in terms of visibility improvement and contrast enhancement. However, ULV fails to correct the color distortion, and the details in some dark regions of MLLE-enhanced results remain imperceptible. On the contrary, our proposed method can produce pleasant results with high contrast and natural appearance.
\begin{figure*}[!t]
\centering
\setlength{\abovecaptionskip}{-0.2cm}
\includegraphics[width=\linewidth]{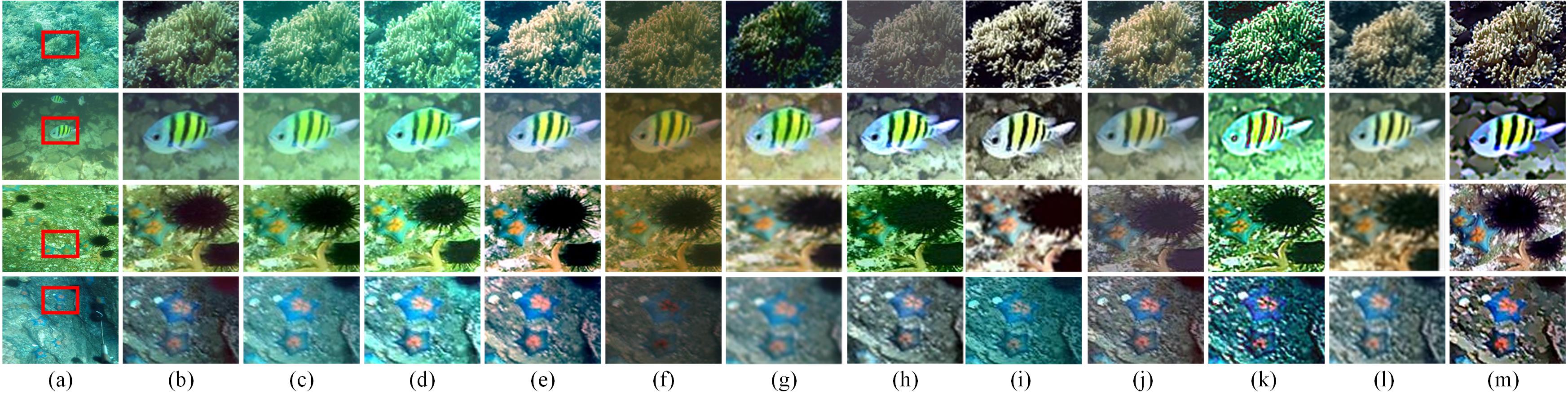}
\caption{Comparisons of local details. (a) the cropped area (red rectangles) from raw images,  the restored results of (b) RCP \cite{31}, (c) IBLA \cite{33}, (d) $ \text{L}^{\text{2}}\text{UWE} $ \cite{05}, (e) SMBLOT \cite{08}, (f) UWCNN \cite{46}, (g) Haze-Lines \cite{09}, (h) BR \cite{27}, (i) MLLE \cite{02}, (j) HLRP \cite{28}, (k) ULV \cite{29}, (l) Semi-UIR \cite{47}, and (m) the proposed method, respectively.}
\label{Fig6}
\end{figure*}
\begin{figure*}[!t]
\centering
\setlength{\abovecaptionskip}{-0.2cm}
\includegraphics[width=\linewidth]{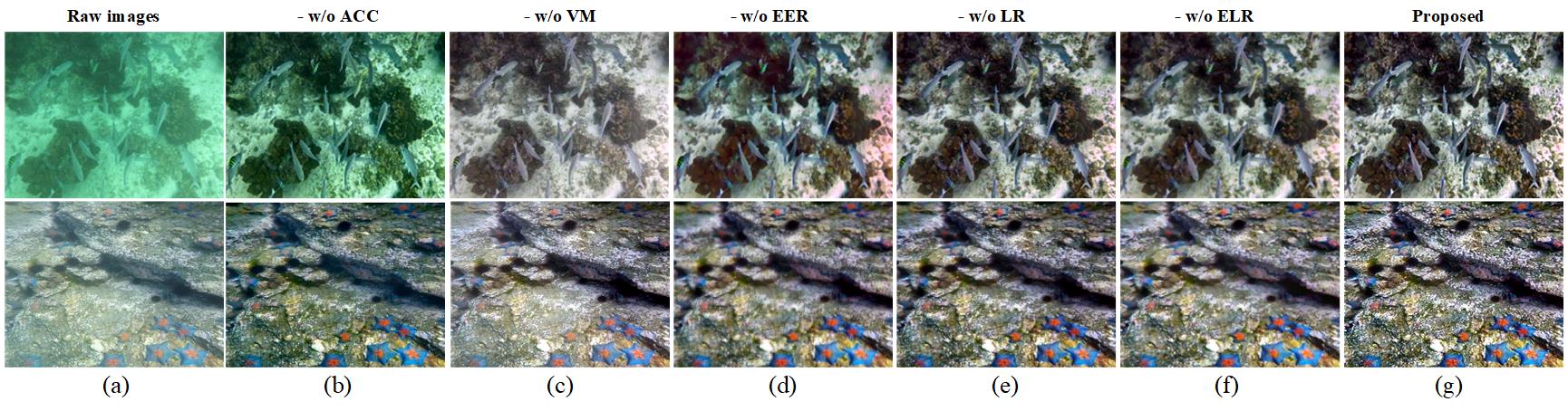}
\caption{Qualitative evaluation of the ablation study. (a) Raw images, (b) the proposed method -w/o ACC, (c) the proposed method -w/o VM, (d) the proposed method -w/o EER, (e) the proposed method -w/o LR, (f) the proposed method -w/o ELR, and (g) the proposed method.}
\label{Fig7}
\end{figure*}

\begin{table*}[!t]
\small
\caption{Quantitative comparisons of FADE and Entropy for the underwater images in Fig. \ref{Fig5}. (The top two performers are highlighted in \textcolor{red}{red} and \textcolor{blue}{blue}, respectively)}
\label{table3}
\centering
\resizebox{\textwidth}{!}{
\begin{tabular}{cccccccccccccccllcccccccc}
\hline
  & \multicolumn{2}{c}{RCP} & \multicolumn{2}{c}{IBLA} & \multicolumn{2}{c}{$ \text{L}^{\text{2}}\text{UWE} $} & \multicolumn{2}{c}{SMBLOT} & \multicolumn{2}{c}{UWCNN} & \multicolumn{2}{c}{Haze-Lines} & \multicolumn{2}{c}{BR} & \multicolumn{2}{l}{MLLE} & \multicolumn{2}{c}{HLRP} & \multicolumn{2}{c}{ULV} & \multicolumn{2}{c}{Semi-UIR} & \multicolumn{2}{c}{Proposed} \\ \cline{2-25} 
   & FADE      & Entropy     & FADE       & Entropy     & FADE       & Entropy      & FADE        & Entropy      & FADE       & Entropy      & FADE          & Entropy        & FADE      & Entropy    & FADE       & Entropy     & FADE       & Entropy     & FADE      & Entropy     & FADE         & Entropy       & FADE         & Entropy       \\ \hline
$\text { (I) }$   & 0.355     & 7.427       & 0.557      & 7.382       & 0.317      & 7.597        & 0.463       & 7.622        & 0.551      & 6.222        & 0.247         &\textcolor{blue}{7.814}          & 0.369     & 7.037      & 0.202      & 7.778       & 0.182      & 7.781       & 0.193     & 7.718       & \textcolor{blue}{0.184}        & 7.767         & \textcolor{red}{0.181}        & \textcolor{red}{7.838}         \\
$\text { (II) }$  & 0.224     & 6.560       & 0.267      & 6.595       & 0.172      & 6.472        & 0.256       & 6.704        & 0.436      & 5.534        & 0.238         & 6.948          & 0.330     & 7.604      & \textcolor{blue}{0.166}      & \textcolor{red}{7.729}       & 0.245      & 7.523       & 0.254     & 5.823       & 0.175        & 7.549         & \textcolor{red}{0.162}        & \textcolor{blue}{7.587}         \\
$\text { (III) }$  & 0.229     & 7.425       & 0.427      & 7.092       & 0.256      & 7.263        & 0.401       & 7.390        & 0.280      & 6.407        & 0.215         & \textcolor{blue}{7.744}         & 0.270     & 7.561      & \textcolor{red}{0.122}      & 7.708     & 0.270      & 7.290       & 0.136     & 7.623       & 0.189        & 7.653         & \textcolor{blue}{0.130}        & \textcolor{red}{7.747}         \\
$\text { (IV) }$  & 0.369     & 6.426       & 0.974      & 6.179       & 0.524      & 6.709        & 0.980       & 6.265        & 0.724      & 6.068        & \textcolor{blue}{0.165}        & 7.292          & 0.216     & \textcolor{blue}{7.473}      & 0.215      & 7.352       & 0.247      & 7.269       & 0.413     & 6.503       & {0.214}        & 7.067         & \textcolor{red}{0.137}        & \textcolor{red}{7.491}         \\
$\text { (V) }$  & 0.295     & 6.417       & 0.383      & 6.373       & 0.232      & 6.090        & 0.331       & 6.890        & 0.773      & 5.276        & 0.165         & 7.179          & 0.215     & \textcolor{blue}{7.382}      & 0.180      & 7.278       & 0.276      & 7.224       & \textcolor{red}{0.144}     & 7.046       & 0.196        & 6.593         & \textcolor{blue}{0.169}        & \textcolor{red}{7.542}         \\
$\text { (VI) }$  & 0.446     & 7.055       & 0.464      & 7.469       & 0.227      & \textcolor{blue}{7.709}        & 0.480       & 7.600        & 0.623      & 6.791        & 0.364         & 6.674          & 0.279     & 7.536      & \textcolor{red}{0.155}      & 7.443       & 0.299      & \textcolor{red}{7.804}       & 0.203     & 7.503       & 0.193        & 7.646         & \textcolor{blue}{0.170}        & 7.595         \\
$\text { (VII) }$  & 0.324     & 7.730       & 0.199      & 7.744       & 0.295      & 7.707        & 0.255       & 7.542        & 0.284      & 6.522        & 0.184         & 7.661          & 0.182     & 7.708      & 0.402      & 7.804       & 0.324      & 7.451       & \textcolor{blue}{0.155}     & \textcolor{red}{7.910}       & 0.185        & 7.815         & \textcolor{red}{0.152}        & \textcolor{blue}{7.824}         \\
$\text { (VIII) }$  & 0.399     & 5.959       & 0.268      & 6.326       & 0.210      & 6.899        & 0.413       & 6.914        & 0.559      & 6.060        & 0.304         & 7.758          & 0.180     & 7.512      & 0.424      & \textcolor{red}{7.842}       & 0.292      & \textcolor{blue}{7.762}       & \textcolor{blue}{0.171}     & 7.049       & 0.334        & 7.242         & \textcolor{red}{0.162}        & 7.611         \\
$\text { (IX) }$  & 0.375     & 7.593       & 0.269      & 7.702       & 0.225      & 7.581        & 0.238       & 7.612        & 0.317      & 6.556        & 0.213         & 7.670          & \textcolor{blue}{0.169}     & 7.614      & 0.363      & \textcolor{blue}{7.754}       & 0.304      & 7.357       & 0.172     & 7.738       & 0.193        & 7.750         & \textcolor{red}{0.141}        & \textcolor{red}{7.788}         \\
$\text { (X) }$ & 0.321     & 7.102       & 0.419      & 7.352       & 0.176      & 7.684        & 0.210       & 7.448        & 0.238      & 7.025        & 0.202         & 7.541          & 0.181     & 7.647      & 0.403      & \textcolor{blue}{7.772}       & 0.229      & 7.657       & \textcolor{red}{0.141}     & 7.745       & 0.264        & 7.670         & \textcolor{blue}{0.159}        & \textcolor{red}{7.811}         \\
$\text { (XI) }$ & 0.313     & 7.003       & 0.974      & 7.284       & 0.524      & 7.636        & 0.980       & 7.736        & 0.249      & 6.970        & 0.159         & 7.682          & 0.165     & 7.633      & 0.316      & \textcolor{blue}{7.864}       & 0.199      & 7.576       & \textcolor{blue}{0.120}     & 7.633       & 0.175        & 7.804         & \textcolor{red}{0.110}        & \textcolor{red}{7.884}         \\
$\text { (XII) }$ & 0.696     & 6.603       & 0.919      & 7.077       & 0.329      & 7.400        & 0.687       & 7.508        & 0.350      & 6.930        & 0.283         & 7.308          & 0.329     & 7.626      & 0.484      & \textcolor{blue}{7.840}       & 0.232      & 7.799       & \textcolor{blue}{0.222}     & 7.253       & 0.343        & 7.683         & \textcolor{red}{0.191}        & \textcolor{red}{7.861}         \\ \hline 
\end{tabular}}
\end{table*}

\begin{table*}[!t]
\small
\caption{Quantitative comparisons of UCIQE and FDUM for the underwater images in Fig. \ref{Fig5}. (The top two performers are highlighted in \textcolor{red}{red} and \textcolor{blue}{blue}, respectively)}
\label{table4}
\centering
\resizebox{\textwidth}{!}{
\begin{tabular}{cccccccccccccccllcccccccc}
\hline
  & \multicolumn{2}{c}{RCP} & \multicolumn{2}{c}{IBLA} & \multicolumn{2}{c}{$ \text{L}^{\text{2}}\text{UWE} $} & \multicolumn{2}{c}{SMBLOT} & \multicolumn{2}{c}{UWCNN} & \multicolumn{2}{c}{Haze-Lines} & \multicolumn{2}{c}{BR} & \multicolumn{2}{l}{MLLE} & \multicolumn{2}{c}{HLRP} & \multicolumn{2}{c}{ULV} & \multicolumn{2}{c}{Semi-UIR} & \multicolumn{2}{c}{Proposed} \\ \cline{2-25} 
   & UCIQE      & FDUM     & UCIQE       & FDUM     & UCIQE       & FDUM      & UCIQE        & FDUM      & UCIQE       & FDUM      & UCIQE          & FDUM        & UCIQE      & FDUM    & UCIQE       & FDUM     & UCIQE       & FDUM     & UCIQE      & FDUM     & UCIQE         & FDUM       & UCIQE         & FDUM       \\ \hline
$\text { (I) }$   & 0.570     & 0.574       & 0.520      & 0.520       & 0.533      & 0.533        & 0.605       & 0.605        & 0.484      & 0.484       & \textcolor{blue}{0.636}         & 0.528          & 0.544     & 0.483      & 0.620      & \textcolor{blue}{0.765}       & 0.583      & 0.583       & 0.597     & 0.597       & 0.614        & 0.705         & \textcolor{red}{0.672}        & \textcolor{red}{0.872}         \\
$\text { (II) }$  & 0.510     & 0.427       & 0.406      & 0.388       & 0.368      & 0.323         & 0.446       & 0.423        & 0.321      & 0.128        & 0.556         & 0.522          & 0.535     & 0.505      & 0.585      & 0.535     
& 0.565      & 0.532       & 0.353     & 0.229       & \textcolor{blue}{0.574}        & \textcolor{blue}{0.567}         & \textcolor{red}{0.622}        & \textcolor{red}{0.692}         \\
$\text { (III) }$  & 0.546     & 0.396       & 0.432      & 0.335       & 0.432      & 0.450         & 0.536       & {0.581}        & 0.452      & 0.310        & \textcolor{blue}{0.599}         & 0.603         & 0.485     & 0.391      & 0.581     & \textcolor{blue}{0.756}       & 0.505      & 0.453       & 0.136     & 0.580       & 0.578        & 0.517         & \textcolor{red}{0.628}        & \textcolor{red}{0.764}         \\
$\text { (IV) }$  & 0.558     & 0.646       & 0.473      & 0.492       & 0.497      & 0.623        & 0.446       & 0.519        & 0.452      & 0.329        & \textcolor{red}{0.702}       & 1.082         &
0.565     & 0.709      & blue{0.652}      & \textcolor{blue}{1.169}       & 0.610      & 0.939       & 0.511     & 0.649       & 0.638        & 0.875         & \textcolor{blue}{0.697}       & \textcolor{red}{1.171}         \\
$\text { (V) }$  & 0.568     & 0.644       & 
0.556      & 0.659       & 0.573      & 0.821        & 0.586       & 0.794        & 0.460      & 0.381        & \textcolor{blue}{0.688}         & 1.110          & 0.573     & 0.738      & 0.646      & \textcolor{blue}{1.275}       & 0.599      & 7.224       & 0.647    & 0.997       & 0.618        & 0.844         & \textcolor{red}{0.735}        & \textcolor{red}{1.293}         \\
$\text { (VI) }$  & 0.568     & 0.432       & 0.565      & 0.448       & 0.584      & 0.663        & 0.576       & 0.398        & 0.489      & 0.307        & \textcolor{blue}{0.641}         & \textcolor{blue}{0.728}          & 0.526     & 0.534      & 0.596      & 0.721       & 0.600      & 0.598       & 0.607     & 0.700       & 0.598        & 0.592        & \textcolor{red}{0.644}        & \textcolor{red}{0.834}         \\
$\text { (VII) }$  & 0.509     & 0.492       & 0.600      & 0.589       & 0.519      & 0.582        & 0.609       & 0.666        & 0.514      & 0.410        & \textcolor{blue}{0.619}         & 0.740          & 0.539     & 0.629      & 0.593      & \textcolor{blue}{0.765}       & 0.498      & 0.408       & 0.607     & 0.712       & 0.593        & 0.595         & \textcolor{red}{0.610}        & \textcolor{red}{0.814}         \\
$\text { (VIII) }$  & 0.401     & 0.344       & 0.429      & 0.377      & 0.432      & 0.438        & 0.552       & 0.578        & 0.469      &
0.317        & \textcolor{blue}{0.623}         & 0.677          & 0.522     & 0.590      & 0.601      & \textcolor{blue}{0.763}       & 0.571      & 0.525       & 0.539     & {0.601}       & 0.523        & 0.536         & \textcolor{red}{0.626}        & \textcolor{red}{0.826}        \\
$\text { (IX) }$  & 0.503     & 0.426       & 0.541      & 0.385       & 0.522      & 0.546        & \textcolor{blue}{0.614}       & 0.674        & 0.456      & 0.367        & 0.612         & \textcolor{blue}{0.775}          & 0.533     & 0.604      & 0.582      & {0.735}       & 0.510      & 0.445       & 0.528     & 0.619       & 0.590        & 0.562         & \textcolor{red}{0.618}        & \textcolor{red}{0.754}         \\
$\text { (X) }$ & 0.490     & 0.342       &
0.556      & 0.488       & 0.533      & 0.585        & 0.563       & 0.582        & 0.511      & 0.349        & \textcolor{red}{0.599}         & 0.670          & 0.535     & 0.519      & 0.570      & \textcolor{blue}{0.673}       & 0.569      & 0.593       & 0.562     & 0.618       & 0.569        & 0.540         & \textcolor{blue}{0.584}        & \textcolor{red}{0.689}         \\
$\text { (XI) }$ & 0.462     & 0.298       &
0.509      & 0.410       & 0.530      & 0.527        & 0.587       & 0.552        & 0.473      & 0.318        & \textcolor{blue}{0.610}         & \textcolor{blue}{0.623}          & 0.523     & 0.486      & 0.583      & 0.550       & 0.549      & 0.518       & 0.569     & {0.567}       & 0.582        & 0.560         & \textcolor{red}{0.615}        & \textcolor{red}{0.699}         \\
$\text { (XII) }$ & 0.393     & 0.168       & 0.467      & 0.294       & 0.455      & 0.355        & 0.500       & 0.336        & 0.462      & 0.259        & 0.614         & 0.518          & 0.495     & 0.287      & 0.568      & 0.511       & \textcolor{red}{0.620}      & \textcolor{red}{0.603}       & 0.488     & 0.405       & 0.554        & 0.408         & \textcolor{blue}{0.573}        & \textcolor{blue}{0.538}         \\ \hline 
\end{tabular}}
\end{table*}

To further illustrate the superiority of our method, we present a detailed comparison of the cropped red rectangle area in Fig. \ref{Fig6}. It can be observed that the restored results of RCP, IBLA, MLLE, and $ \text{L}^{\text{2}}\text{UWE} $ still exhibit residual fog. Likewise, UWCNN, Semi-UIR, and BR methods appear to destroy the structural information in the image. SMBLOT, Haze-Lines, and HLRP have a slight effect on recovering the content details. At first glance, the outcomes produced by ULV show more details. But after careful inspection, we can see that the ULV brings red artifacts around the edges of the image. Compared with these UIER methods, our method can reveal more  details and edge information, leading to significant improvement in
sharpness.

The quantitative results associated to Fig. \ref{Fig5} are reported in Tables \ref{table3}-\ref{table4}. It can be easily observed that the proposed method performs best for most cases in terms of FADE, Entropy, UCIQE, and FDUM. Specifically. As shown in Table \ref{table3}, our method obtains almost all the best FADE values, which proves its superior capability in dehazing. Regarding to Entropy, MLLE and our method rank higher than other compared methods, because they both can efficiently enhance the contrast of images. With respect to UCIQE, the score distribution of other methods is between 0.3 and 0.6, while the score of the proposed method is stable above 0.57, indicating that the restored images by our method have a better balance among saturation, contrast, and chromaticity. Also, for FDUM, our method acquires the best results in most cases except \text {XII} generated by HLRP. Nevertheless, combined with the qualitative assessment in Fig. \ref{Fig5}, the corresponding results enhanced by HLRP have a low contrast. Overall, the restored results show that our method has better robustness across all degraded images. 

Moreover, we conduct a comprehensive quantitative comparison on the entire UIEB and UIQS datasets, as presented in Table \ref{Table5}. Our method achieves the best average values of FADE, Entropy, and UCIQE on the UIEB dataset, as well as the best average values of Entropy and UCIQE on the UIQS dataset. Although our method ranks second in FADE on the UIQS dataset, the difference from the top-ranked method is minimal. In addition, our method achieves competitive performance against the SOTA methods in terms of the FDUM metric. In summary, the qualitative and quantitative comparisons both demonstrate the superior performance of our proposed method compared to several SOTA methods.

\begin{table} [!t]
\setlength{\abovecaptionskip}{-0.2cm}  %段前
\small
\centering
\caption{Quantitative comparisons of average values of FADE, Entropy, UCIQE, and FDUM on the UIEB and UIQS datasets. (The top two   performers are highlighted in \textcolor{red}{red} and \textcolor{blue}{blue}, respectively.)}
\label{Table5}
\centering
\scalebox{0.7}{
\begin{tabular}{ccccc|cccc} \\
\hline
\multicolumn{1}{c}{\multirow{2}{*}{Methods}} & \multicolumn{4}{c|}{\multirow{1}{*}{UIEB}} & \multicolumn{4}{c}{\multirow{1}{*}{UIQS}}    \\ \cline{2-9} 
 & FADE & Entropy  & UCIQE & FDUM   & FADE & Entropy  & UCIQE & FDUM  \\ \hline
RCP    &0.432	&7.181	&0.581	&0.562	&0.414	&6.934	&0.472	&0.316\\
IBLA   &0.435	&7.143	&0.564	&0.579	&0.387	&7.205	&0.521	&0.403 \\
$ \text{L}^{\text{2}}\text{UWE} $ &0.287	&7.337	&0.559	&0.714	&0.237	&7.448	&0.513	&0.512 \\
SMBLOT  	&0.436	&7.338	&0.606	&0.721	&0.426	&7.508	&0.580	&0.571 \\
UWCNN   &0.630	&6.441	&0.483	&0.366	&0.457	&6.892	&0.484	&0.310 \\
Haze-Lines   &0.309	&7.373	&\textcolor{blue}{0.642}	& {0.814}	&0.270	&7.476	&\textcolor{blue}{0.596}	&\textcolor{red}{0.700} \\
BR      &0.308	&7.522	&0.565	&0.662	&0.238	&7.603	&0.527	&0.508 \\
MLLE      &0.297	& \textcolor{blue}{7.625}	&0.607	& \textcolor{red}{0.889}	&0.262	&\textcolor{blue}{7.715}	&0.577	&0.636 \\
HLRP  & 0.302	&7.277	&{0.616}	&0.823	&0.280	&7.524	&0.579	&{0.614} \\
ULV       &\textcolor{blue}{0.229}	& 7.386	&{0.611}	&0.826	&\textcolor{red}{0.185}	&7.395	&{0.558}	&0.588 \\
Semi-UIR       &0.304	&{7.516}	&{0.615}	&0.690	&{0.306}	&7.678	&0.575	&0.523 \\
Proposed    & \textcolor{red}{0.219} & \textcolor{red}{7.634} & \textcolor{red}{0.648} & \textcolor{blue}{0.871} & \textcolor{blue}{0.190} & \textcolor{red}{7.741} & \textcolor{red}{0.607} & \textcolor{blue}{0.679}  \\ \hline
\end{tabular}}
\end{table}

\begin{table} [!t]
\setlength{\abovecaptionskip}{-0.2cm}  %段前
\small
\centering
\caption{Quantitative evaluation of ablation study on the UIEB and UIQS datasets. (The top two   performers are highlighted in \textcolor{red}{red} and \textcolor{blue}{blue}, respectively.)}
\label{Table6}
\centering
\scalebox{0.7}{
\begin{tabular}{ccccc|cccc} \\
\hline
\multicolumn{1}{c}{\multirow{2}{*}{Methods}} & \multicolumn{4}{c|}{\multirow{1}{*}{UIEB}} & \multicolumn{4}{c}{\multirow{1}{*}{UIQS}}    \\ \cline{2-9} 
 & FADE & Entropy  & UCIQE & FDUM   & FADE & Entropy  & UCIQE & FDUM  \\ \hline
-w/o ACC    & \textcolor{blue}{0.232}	&7.092	&0.564	&0.691	& \textcolor{blue}{0.221}	&7.162	&0.508	&0.578\\
-w/o VM    &0.407	&7.433	&0.584	&0.595	&0.391	&7.460	&0.563	&0.469 \\
w/o EER  	&0.273	&7.530	& \textcolor{blue}{0.637}	&0.685	&0.258	&7.545	& 0.588&0.559 \\
-w/o LR    &0.266	&\textcolor{blue}{7.613}	& 0.623	& \textcolor{blue}{0.700}	&0.229	& \textcolor{blue}{7.671}	&\textcolor{blue}{0.594}	& \textcolor{blue}{0.582} \\
-w/o ELR   &0.281	&7.598	&0.592	&0.663	&0.246	&7.636	&0.579	&0.564 \\
Proposed    & \textcolor{red}{0.219} & \textcolor{red}{7.634} & \textcolor{red}{0.648} & \textcolor{red}{0.871} & \textcolor{red}{0.190} & \textcolor{red}{7.741} & \textcolor{red}{0.607} & \textcolor{red}{0.679}  \\ \hline
\end{tabular}}
\end{table}
\vspace{-3mm}
\subsection{Ablation Studies}
To demonstrate the effectiveness of each key component of the proposed method, we further perform some ablation studies on the UIEB and UIQS datasets, including, Fig. \ref{Fig7}(b) the proposed method without adaptive color correction (-w/o ACC), Fig. \ref{Fig7}(c) the proposed method without variation model (-w/o VM), Fig. \ref{Fig7}(d) the proposed method without Euler’s elastica (-w/o EER), Fig. \ref{Fig7}(e) the proposed method without Laplacian regularization (-w/o LR), Fig. \ref{Fig7}(f) the proposed method without Euler’s elastica and Laplacian regularization (-w/o ELR), and Fig. \ref{Fig7}(g) the proposed method. From Fig. \ref{Fig7}(b), we can observe that our -w/o ACC has a remarkable performance in removing the haze-like appearance of underwater images, but cannot effectively remove color deviation. On the other hand, as shown in Fig. \ref{Fig7}(c), the color cast can be removed to a certain extent by our -w/o VM, but it does not work well for improving the contrast of the image. Referring to Fig. \ref{Fig7}(d), the local details of our -w/o EER are unsatisfactory. It can be seen from Fig. \ref{Fig7}(e) that the perception quality of our -w/o LR has been significantly improved, but there is still room for contrast enhancement. In Fig. \ref{Fig7}(f), the result of our -w/o ELR seems a little blurry. By contrast, the restored underwater images by the proposed method are characterized by higher contrast and more valuable details.

In addition, the quantitative evaluations of the ablation study conducted on the whole UIEB and UIQS  datasets are tabulated in Table \ref{Table6}. It can be seen that the proposed method achieves the best values in terms of FADE, Entropy, UCIQE, and FDUM metrics, which also demonstrates that each component has a positive impact on the proposed method.

\subsection{Potential Applications}
Although the proposed method is specially designed for recovering underwater images, it can also be well generalized to other visual tasks. To demonstrate the  broad applicability of the proposed method, we further conduct three other applications including outdoor image defogging, low-light image restoration, and feature points matching. The compared methods include both underwater and outdoor image visual enhancement ones: GDCP \cite{55}, $ \text{L}^{\text{2}}\text{UWE} $ \cite{05}, CCDID  \cite{56}, SMBLOT \cite{08}, BR \cite{23}, HLRP \cite{25}, Semi-UIR \cite{47} and SPCV \cite{57}.
\begin{figure*}[!t]
\centering
\setlength{\abovecaptionskip}{-0.2cm}
\includegraphics[width=\linewidth]{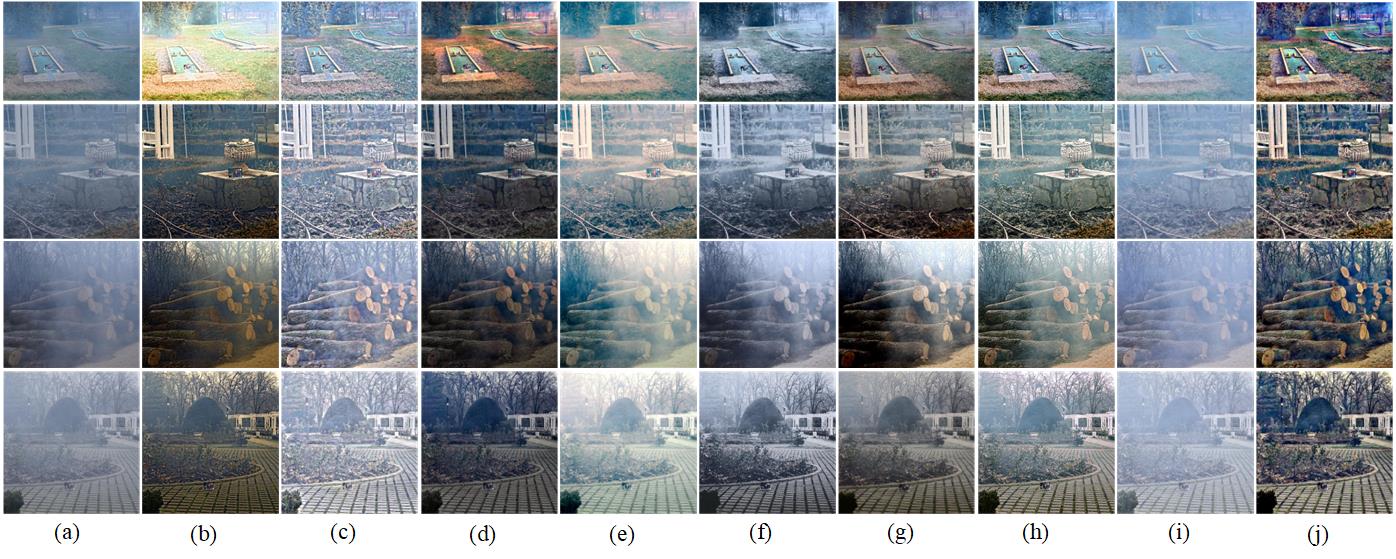}
\caption{Qualitative comparison of outdoor foggy images. (a) Raw images, the restored results of (b) GDCP \cite{55}, (c) $ \text{L}^{\text{2}}\text{UWE} $ \cite{05}, (d) CCDID \cite{56}, (e) SMBLOT \cite{08}, (f) BR \cite{27}, (g) HLRP \cite{28}, (h) Semi-UIR \cite{47} and (i) SPCV \cite{57}, and (j) the proposed method, respectively.}
\label{Fig8}
\end{figure*}
\begin{figure*}[!t]
\centering
\setlength{\abovecaptionskip}{-0.2cm}
\includegraphics[scale=0.4]{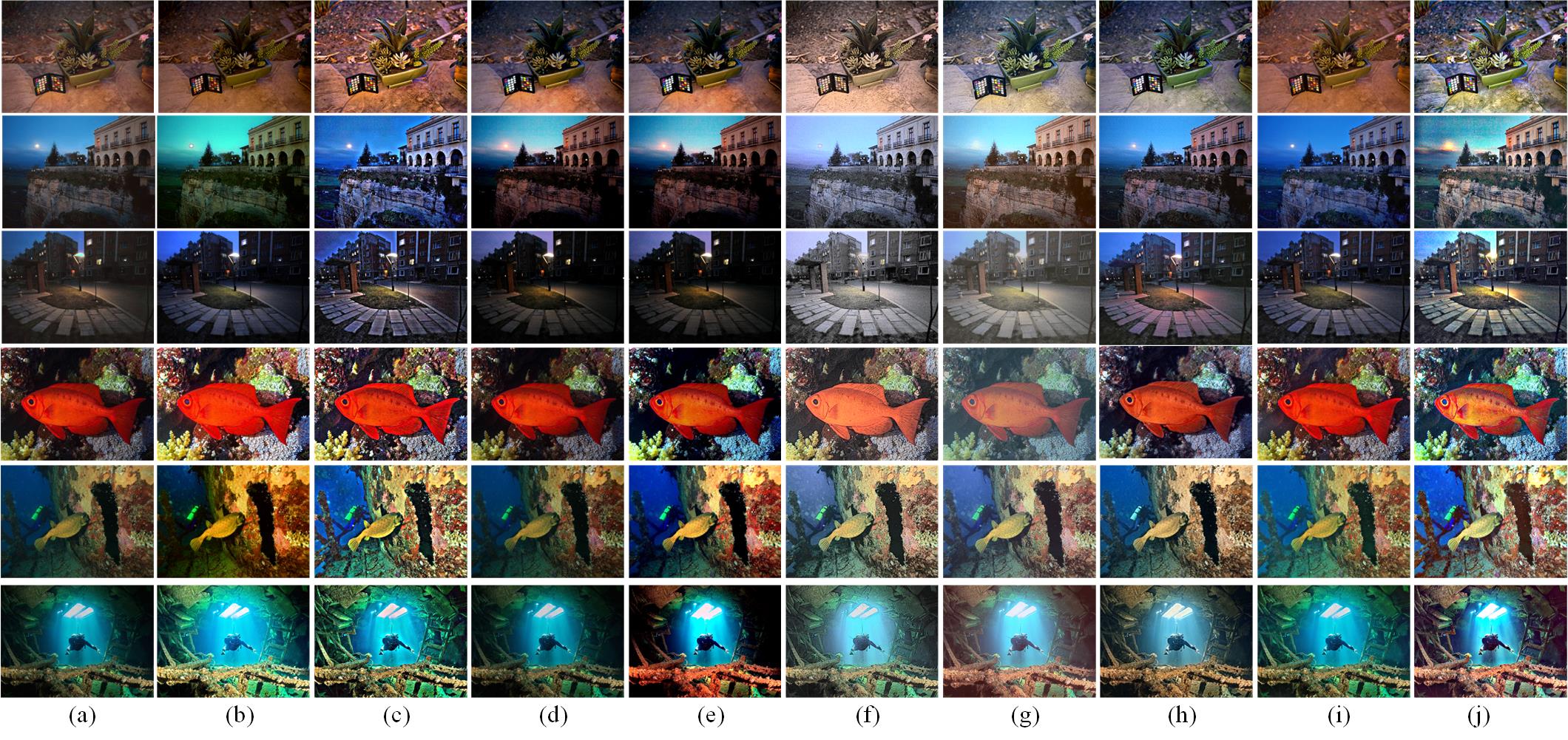}
\caption{Qualitative comparison of low-light images. (a) Raw images, the restored results of (b) GDCP \cite{55}, (c) $ \text{L}^{\text{2}}\text{UWE} $ \cite{05}, (d) CCDID  \cite{56}, (e) SMBLOT \cite{08}, (f) BR \cite{27}, (g) HLRP \cite{28}, (h) Semi-UIR \cite{47} and (i) SPCV \cite{57}, and (j) the proposed method, respectively.}
\label{Fig9}
\end{figure*}

\textbf{Outdoor image dehazing.} Since the underwater imaging model is similar to the outdoor fogging model. Therefore, we first apply the proposed method to restore images captured under foggy weather. The enhanced results of different methods on outdoor foggy images are shown in Fig. \ref{Fig8}. We can observe that the GDCP and CCDID  methods can remove the haze, but it is still challenging to unveil more details. Although $ \text{L}^{\text{2}}\text{UWE} $ shows a good performance in contrast enhancement, the dehazing effect is relatively poor. Intuitively, the enhanced results of SMBLOT and Semi-UIR tend to produce some color deviation. In comparison, BR improves the visibility of fog images, while it also produces partial darkness. Similarly, both HLRP and SPCV methods are unsatisfactory in dehazing. On the contrary, the proposed method comprehensively
outperforms the above methods in terms of dehazing, contrast
enhancement and texture revealing.
\begin{figure}[!t]
\centering
\setlength{\abovecaptionskip}{-0.2cm}
\includegraphics[width=7.5cm,height=4cm]{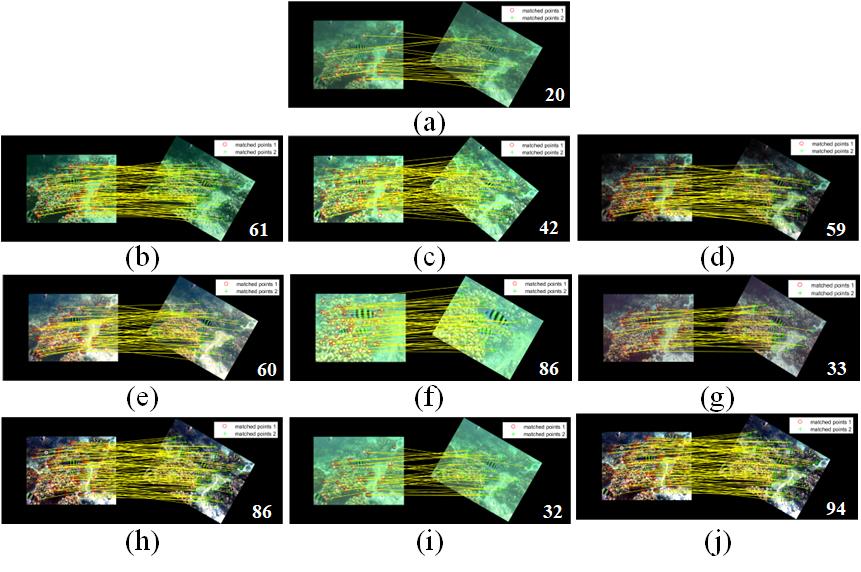}
\caption{Application test using SURF. (a) Raw images, the corresponding results obtained by (b) GDCP \cite{55}, (c) $ \text{L}^{\text{2}}\text{UWE} $ \cite{05}, (d) CCDID  \cite{56}, (e) SMBLOT \cite{08}, (f) BR \cite{27}, (g) HLRP \cite{28}, (h) Semi-UIR \cite{47} and (i) SPCV \cite{57}, and (j) the proposed method, respectively.}
\label{Fig10}
\end{figure}

\textbf{ Low-light image restoration.} Outdoor and underwater images captured under low-light conditions are seriously degraded with low visibility and high-level noise. In this part, we also examine the ability of our proposed method to restore low-light images, as shown in Fig. \ref{Fig9}. We can easily find that although the GDCP is effective in brightness adjustment of low-light underwater images, it has an inappreciable effect on low-light outdoor images. Similarly, in spite of the good performance in brightness adjustment, $ \text{L}^{\text{2}}\text{UWE} $ is prone to produce images with an unnatural appearance. CCDID  and ULV introduce some invisible dark areas. Likewise, BR, HLRP, and Semi-UIR can improve the illumination and visibility of low-light images, but they also blur the details. The recovered images produced by SPCV present better naturalness, unfortunately, it is still challenging to adequately expose the dark regions. In contrast, our method can not only improve the brightness of the image but also preserve more details, thereby generating some results with a more realistic appearance.
\vspace{8mm}

\textbf{ Local feature points matching.} To further validate the effectiveness of the proposed method in information recovery, the number of matching feature points in the SURF \cite{58} is counted on the original underwater images and their restored results of different compared methods, as shown in Fig. \ref{Fig10}. It can be clearly observed that our method can significantly increase the number of local feature point matches compared with other advanced methods. Therefore, it can be concluded that the proposed method recovers more crucial features, which is beneficial to high-level visual applications.
\section{CONCLUSION}\label{CONCLUSION}
In this paper, we introduce a novel dual high-order regularization-guided variational framework for underwater image restoration. Unlike most existing model-based methods that ignore the local illumination variations present in a single image, we establish an extended underwater imaging model by taking local illumination differences into account. In our proposed framework, we first utilize the weight factors-based color compensation to compensate the attenuated color channels and then remove color cast by a color balance strategy. Furthermore, the local brightest pixel estimation and an improved gamma correction strategy are combined to acquire the local ambient illuminance. Also, a fast optimization algorithm based on ADMM is designed to accelerate  the proposed variational model. Comprehensive qualitative and quantitative experiments reveal that our method achieves superior performance in color correction, dehazing, and structure preservation against other state-of-the-art methods.

\appendix
\section{Appendix}
\label{Appendix}

In this appendix, we present a detailed procedure utilizing the ADMM algorithm to derive the solutions for the minimization problems (P1), (P2), and (P3). 

 \textbf{Update for P1:}  $R$ and $ L$ Reconstruction.

 $R$ and $L$ sub-equations: The $R$ and $L$ subproblems involve quadratic optimization problems. By calculating (\ref{12-a}) and (\ref{12-b}), we can formulate their corresponding Euler-Lagrange equations for the iteration loop indexed at k+1 as: 

\begin{subequations}
\begin{align}
&\left\{\begin{aligned}
&R^{k+1}\left(L^{k} \mathrm{t}\right)^{2}-f-\mu_{1} \Delta R^{k+1} \quad x \in \Omega \\
&\mu_{1} \nabla \cdot \vec{w}^{k}+\nabla \cdot \vec{\lambda}_{1}^{k}=0 \\
&\left(\nabla R^{k+1}-\vec{w}^{k}-\frac{\vec{\lambda}_{1}^{k}}{\mu_{1}}\right) \cdot \vec{n}=0 \quad x \in \partial \Omega
\end{aligned},\right.
\tag{15-a}
\label{15-a}\\
&\left\{\begin{array}{l}
\left(R^{k+1} \mathrm{t}+1-t\right)^{2} L^{k+1}-h-\mu_{5} \Delta L^{k+1} \quad x \in \Omega \\
+\mu_{1} \nabla \cdot \vec{m}^{k}+\nabla \cdot \vec{\lambda}_{5}^{k}=0 \\
\left(\nabla L^{k+1}-\vec{m}^{k}-\frac{\vec{\lambda}_{5}^{k}}{\mu_{5}}\right) \cdot \vec{n}=0 \quad x \in \partial \Omega
\end{array},\right.
\tag{15-b}
\label{15-b}
\end{align}
\end{subequations}
where $ f=L^{k} I t-L^{k} t \cdot(1-t) L^{k} $, $ \vec{n} $ is a normal vector, $ \Delta R^{k+1}=R_{i+1, j}^{k}+R_{i-1, j}^{k}+R_{i, j+1}^{k}, R_{i, j-1}^{k}-4 R_{i, j}^{k+1} $, $ h=I^{k}\left(R^{k+1} \mathrm{t}+1-t\right) $, $ \Delta L^{k+1}=L_{i+1, j}^{k}+L_{i-1, j}^{k}+L_{i, j+1}^{k}, L_{i, j-1}^{k}-4 L_{i, j}^{k+1} $.

\noindent Next, we can get $R^{k+1}$ and $L^{k+1}$ by discretization:
\begin{subequations}
\begin{align}
R^{k+1}=\frac{f+\mu_{1} \Delta R^{k+1}-\mu_{1} \nabla \cdot \vec{w}^{k}-\nabla \cdot \vec{\lambda}_{1}^{k}}{\left(L^{k+1} t\right)^{2}+4 \mu_{1}},
\tag{16-a}
\label{16-a}\\
L^{k+1}=\frac{h+\mu_{5} \Delta L^{k+1}-\mu_{5} \nabla \cdot \vec{m}^{k}-\nabla \cdot \vec{\lambda}_{5}^{k}}{\left(R^{k+1} t+1-t\right)^{2}+4 \mu_{5}} .
\tag{16-b}
\label{16-b}
\end{align}
\end{subequations}

\textbf{Update for P2:}  Derivation of Auxiliary Variables for $R$ Reconstruction.

  $\vec{w}$ sub-equation: The solution of the variable $\vec{w}$ is a non-convex problem, which can be acquired using the soft threshold operation. The minimizer $\vec{w}$ is derived by:
\begin{subequations}
\begin{align}
\vec{w}_{i, j}^{k+1}=\max \left(\left|\vec{A}_{i, j}\right|-\frac{B_{i, j}}{\mu_{1}}, 0\right) \frac{\vec{A}_{i, j}}{\left|\vec{A}_{i, j}\right|}, 0 \frac{\overrightarrow{0}}{|\overrightarrow{0}|}=\overrightarrow{0},
\tag{17}
\label{17}
 \end{align}
\end{subequations}
where $\vec{A}_{i, j}=\nabla R_{i, j}^{k+1}+\frac{\left(\lambda_{2 i, j}^{k}+\mu_{2}\right) \vec{q}_{i, j}^{k}-\vec{\lambda}_{1 i,j}^{k}}{\mu_{1}}$, $ B_{i, j}=\lambda_{2 i, j}^{k}+\mu_{2}+\alpha+\beta\left(v_{i, j}^{k}\right)^{2} $. 

 $ \vec{p} $ sub-equation: According to (\ref{13-b}), the Euler-Lagrange equation of $ \vec{p} $ can be easy to obtain as follows:
\begin{subequations}
\begin{align}
\left\{\begin{array}{cc}\mu_{4}\left(\vec{p}^{k+1}-\vec{q}^{k}+\frac{\vec{\lambda}_{4}^{k}}{\mu_{4}}\right)-\mu_{3}\\ \nabla\left(\nabla \cdot \vec{p}^{k+1}-v^{k+1}-\frac{\lambda_{3}^{k}}{\mu_{3}}\right)=0 & x \in \Omega \\\left(\nabla \cdot \vec{p}^{k+1}-v^{k+1}-\frac{\lambda_{3}^{k}}{\mu_{3}}\right) \cdot \vec{n}=0 & x \in \partial \Omega\end{array} ,\right.
\tag{18}
\label{18}
 \end{align}
\end{subequations}
then, the solution of $ \vec{p} $ is updated by the fast Fourier transform (FFT):
\begin{subequations}
\begin{align}
\left\{\begin{array}{l}\vec{p}_{1 i, j}^{k+1}=\operatorname{Re}\left(FFT^{-1}\left(\frac{a_{22} FFT\left(h_{1 i, j}^{k}\right)-a_{12} FFT\left(h_{2 i, j}^{k}\right)}{\left(\frac{\mu_{4}}{\mu_{3}}\right)^{2}-\frac{2 \mu_{4}}{\mu_{3}}\left(\cos Z_{i}+\cos Z_{j}-2\right)}\right)\right) \\\vec{p}_{2 i, j}^{k+1}=\operatorname{Re}\left(FFT^{-1}\left(\frac{a_{11} FFT\left(h_{2 i, j}^{k}\right)-a_{21} FFT\left(h_{1 i, j}^{k}\right)}{\left(\frac{\mu_{4}}{\mu_{3}}\right)^{2}-\frac{2 \mu_{4}}{\mu_{3}}\left(\cos Z_{i}+\cos Z_{j}-2\right)}\right)\right)\end{array} ,\right.
\tag{19}
\label{19}
 \end{align}
\end{subequations}
where $ h_{1 i, j}^{k}=-\partial_{x} v_{i,j}^{k}-\frac{\partial_{x} \lambda_{3_{1i,j}}^{k}}{\mu_{3}}+\frac{\mu_{4}}{\mu_{3}} q_{1 i,j}-\frac{\lambda_{4_{1i,j}}^{k}}{\mu_{3}} $, $ h_{2 i, j}^{k}=-\partial_{y} v_{i,j}^{k}-\frac{\partial_{y} \lambda_{3_{2i,j}}^{k}}{\mu_{3}}+\frac{\mu_{4}}{\mu_{3}} q_{2 i,j}-\frac{\lambda_{4_{2i,j}}^{k}}{\mu_{3}} $, the coefficients are: 
\begin{subequations}
\begin{align}
\left\{\begin{array}{l}
a_{11}=\frac{\mu_{4}}{\mu_{3}}-2\left(\cos Z_{i}-1\right) \\
a_{12}=-\left(1-\cos Z_{j}+\sqrt{-1} \sin Z_{j}\right)\\
\left(-1+\cos Z_{i}+\sqrt{-1} \sin Z_{i}\right) \\
a_{21}=-\left(1-\cos Z_{i}+\sqrt{-1} \sin Z_{i}\right)\\
\left(-1+\cos Z_{j}+\sqrt{-1} \sin Z_{j}\right) \\
a_{22}=\frac{\mu_{4}}{\mu_{3}}-2\left(\cos Z_{j}-1\right)
\end{array} \right.
\tag{20}
\label{20}
\end{align}
\end{subequations}
Here, $Z_{i}=\left(2 \pi / N_{1}\right) \omega_{i}, \omega_{i}=1, \ldots, N_{1}$ and $Z_{j}=\left(2 \pi / N_{2}\right) \omega_{j}, \omega_{j}=1, \ldots, N_{2}$. $N_{1}$ and $N_{2}$ are the number of pixels.

$ \vec{q} \text { and } v $ sub-equations: The Euler-Lagrange equations with respect to $ \vec{q} $ and $ v $ can be expressed as follows:
\begin{subequations}
\begin{align}
\overrightarrow{\tilde{q}}^{k+1}=\vec{p}^{k+1}-\frac{\vec{\lambda}_{4}^{k}-\left(\lambda_{2}^{k}+\mu_{2}\right) \vec{w}^{k+1}}{\mu_{4}}=0,
\tag{21-a}
\label{21-a}\\
2b\left|\vec{w}^{k+1}\right| v^{k+1}+\mu_{3}\left(v^{k+1}-\nabla \cdot \vec{p}^{k+1}+\frac{\lambda_{3}}{\mu_{3}}\right)=0  ,
\tag{21-b}
\label{21-b}
\end{align}
\end{subequations}
the solutions of $ \vec{q} \text { and } v $ are easily obtained as follows:
\begin{subequations}
\begin{align}
\vec{q}^{k+1}=\frac{\overrightarrow{\tilde{q}}^{k+1}}{\max \left(1,\left|\overrightarrow{\tilde{q}}^{k+1}\right|\right)},
\tag{22-a}
\label{22-a}\\
v^{k+1}=\frac{\left(\mu_{3} \nabla \cdot \vec{p}^{k+1}-\lambda_{3}\right)}{\mu_{3}+2 \beta\left|\vec{w}^{k+1}\right|} .
\tag{22-b}
\label{22-b}
\end{align}
\end{subequations}

 \textbf{Update for P3:} Derivation of Auxiliary Variables for $L$ Reconstruction.

$ \vec{m} $ sub-equation: During the iterative process for variable $ \vec{m} $ , we still employ the FFT algorithm :
\begin{subequations}
\begin{align}
\left\{\begin{array}{l}
\vec{m}_{1 i, j}^{k+1}=\operatorname{Re}\left(F F T^{-1}\left(\frac{a_{22} F F T\left(n_{1 i, j}^{k}\right)-a_{m 12} F F T\left(n_{2 i, j}^{k}\right)}{\left(\frac{\mu_{5}}{\mu_{6}}\right)^{2}-\frac{2 \mu_{5}}{\mu_{6}}\left(\cos Z_{i}+\cos Z_{j}-2\right)}\right)\right. \\
\vec{m}_{2 i, j}^{k+1}=\operatorname{Re}\left(F F T^{-1}\left(\frac{a_{11} F F T\left(n_{1 i, j}^{k}\right)-a_{m 21} F F T\left(n_{2 i, j}^{k}\right)}{\left(\frac{\mu_{5}}{\mu_{6}}\right)^{2}-\frac{2 \mu_{5}}{\mu_{6}}\left(\cos Z_{i}+\cos Z_{j}-2\right)}\right)\right.
\end{array},\right.
\tag{23}
\label{23}
 \end{align}
\end{subequations}
where $a_{m 21}=\frac{\mu_{6}}{\mu_{5}}-2\left(\cos Z_{i}-1\right), a_{m 22}=\frac{\mu_{6}}{\mu_{5}}-2\left(\cos Z_{j}-1\right), n_{1 i, j}^{k}=-\partial_{x} g_{i, j}^{k}-\frac{\partial_{x} \lambda_{61 i, j}^{k}}{\mu_{6}}+\frac{\mu_{5}}{\mu_{6}} \partial_{x} L-\frac{\lambda_{51 i, j}^{k}}{\mu_{6}}, n_{2 i, j}^{k}=-\partial_{y} g_{i, j}^{k}-\frac{\partial_{y} \lambda_{\theta_{i, j}}^{k}}{\mu_{6}}+\frac{\mu_{5}}{\mu_{6}} \partial_{y} L-\frac{\lambda_{5_{i, j}}^{k}}{\mu_{6}}.$ 

$ g $ sub-equation: The Euler-Lagrange equations of $ g $ is a linear equation:
\begin{subequations}
\begin{align}
\gamma \cdot g^{k+1}+\lambda_{6}^{k} g^{k+1}+\mu_{6} \cdot\left(g^{k+1}-\nabla m^{k+1}\right)=0 ,
\tag{24}
\label{24}
\end{align}
\end{subequations}
with a closed form solution:
\begin{subequations}
\begin{align}
g^{k+1}=\frac{\mu_{6} \cdot \nabla m^{k+1}-\lambda_{6}^{k}}{\mu_{6}+\gamma} .
\tag{25}
\label{25}
\end{align}
\end{subequations}

 \noindent\textbf{Updating Lagrange multipliers:}
\begin{subequations}
\begin{align}
\left\{\begin{array}{l}\vec{\lambda}_{1}^{k+1}=\vec{\lambda}_{1}^{k}+\mu_{1}\left(\vec{w}^{k+1}-\nabla R^{k+1}\right) \\\lambda_{2}^{k+1}=\lambda_{2}^{k}+\mu_{2}\left(\left|\vec{w}^{k+1}\right|-\vec{w}^{k+1} \cdot \vec{q}^{k+1}\right) \\\lambda_{3}^{k+1}=\lambda_{3}^{k}+\mu_{3}\left(v^{k+1}-\nabla \cdot \vec{p}^{k+1}\right) \\\vec{\lambda}_{4}^{k+1}=\vec{\lambda}_{4}^{k}+\mu_{4}\left(\vec{p}^{k+1}-\vec{q}^{k+1}\right) \\\vec{\lambda}_{5}^{k+1}=\vec{\lambda}_{5}^{k}+\mu_{5}\left(\vec{m}^{k+1}-\nabla L^{k+1}\right) \\\lambda_{6}^{k+1}=\lambda_{6}^{k+1}+\mu_{6}\left(g^{k+1}-\nabla \cdot \vec{m}^{k+1}\right)\end{array} .\right.
\tag{26}
\label{26}
\end{align}
\end{subequations}

\end{document}